\documentclass{article}

\PassOptionsToPackage{numbers, compress}{natbib}
\usepackage[eandd, final]{neurips_2026}

\usepackage{microtype}
\usepackage{graphicx}
\graphicspath{{./plots/}}
\usepackage{subcaption}
\usepackage{booktabs}
\usepackage{hyperref}
\usepackage{xcolor}

\usepackage{amsmath}
\usepackage{amssymb}
\usepackage{mathtools}
\usepackage{amsthm}

\usepackage[capitalize,noabbrev]{cleveref}

\usepackage{siunitx}
\usepackage{xspace}

\usepackage[textsize=tiny]{todonotes}

\usepackage{listings}
\theoremstyle{plain}

\theoremstyle{definition}

\theoremstyle{remark}

\newcommand{\MAT}[1]{\ensuremath{\mathbf{#1}}}

\newcommand{\TT}{\ensuremath{^{\mathrm{T}}}}

\newcommand{\OP}[1]{\operatorname{#1}}

\newcommand{\EQPUNCTUATIONSTYLE}[1]{\textrm{#1}}
\newcommand{\EQDOT}{\,\EQPUNCTUATIONSTYLE{.}}
\newcommand{\EQCOMMA}{\,\EQPUNCTUATIONSTYLE{,}}

\NewDocumentCommand\NORM{om}{
  \IfNoValueTF{#1}
    {\ensuremath{\lVert #2 \rVert}}
    {\ensuremath{\lVert #2 \rVert_{#1}}}
}

\NewDocumentCommand\ABS{om}{
  \IfNoValueTF{#1}
    {\ensuremath{\lvert #2 \rvert}}
    {\ensuremath{\lvert #2 \rvert_{#1}}}
}

\NewDocumentCommand\EXPVAL{om}{
  \IfNoValueTF{#1}
    {\ensuremath{\mathbb{E}\!\left[#2\right]}}
    {\ensuremath{\mathbb{E}_{#1}\!\left[#2\right]}}
}

\NewDocumentCommand\IND{d<>o}{
  \IfNoValueTF{#1}
    {\IfNoValueTF{#2}{}{\ensuremath{_{#2}}}}
    {\IfNoValueTF{#2}
      {\ensuremath{_{\textrm{#1}}}}
      {\ensuremath{_{\textrm{#1}, #2}}}}
}
\NewDocumentCommand\SUP{d<>o}{
  \IfNoValueTF{#1}
    {\IfNoValueTF{#2}{}{\ensuremath{^{\left(#2\right)}}}}
    {\IfNoValueTF{#2}
      {\ensuremath{^{\textrm{#1}}}}
      {\ensuremath{^{\textrm{#1}, \left(#2\right)}}}}
}

\NewDocumentCommand\MIN{o}{
  \IfNoValueTF{#1}{\min}{\min\limits_{#1}\,}
}
\NewDocumentCommand\MAX{o}{
  \IfNoValueTF{#1}{\max}{\max\limits_{#1}\,}
}

\newcommand\CDOT{{\mkern 1mu\cdot\mkern 1mu}}
\newcommand{\TIMES}{{\times}}

\makeatletter
\DeclareRobustCommand\onedot{\futurelet\@let@token\@onedot}
\def\@onedot{\ifx\@let@token.\else.\null\fi\xspace}
\makeatother
\newcommand{\custompar}[1]{\noindent\textbf{#1:\;}}

\newcommand{\LatinStyle}{\textit}

\newcommand\eg{\LatinStyle{e}.\LatinStyle{g}\onedot}

\newcommand\ie{\LatinStyle{i}.\LatinStyle{e}\onedot}

\newcommand\wrt{w.\kern0.05em r.\kern0.05em t\onedot}

\newcommand{\bigO}[1]{\ensuremath{\mathcal{O}\!\left(#1\right)}}

\newcommand{\URL}[1]{\url{#1}}
\newcommand{\REPOURL}{\URL{https://github.com/SAP-samples/tabular-attention-benchmark}}
\newcommand{\REPOURLSHORT}{\href{https://github.com/SAP-samples/tabular-attention-benchmark}{https://github.com/SAP-samples/tabular-attention-benchmark}}

\title{Benchmarking Attention for\\ Tabular Foundation Models}

\author{%
Maximilian Schambach\;\quad%
Clemens Biehl\;\quad%
Sam Thelin\\[2mm]
SAP SE\\
\texttt{\{firstname.lastname\}@sap.com}
}

\begin{document}

\maketitle

\begin{abstract}
Tabular in-context learners such as TabPFN, Mitra, or ConTextTab rely on alternating row and column attention over 2D sequences of latent embeddings.
These attention patterns differ markedly from the one-dimensional case in language models: row attention involves longer sequences while column attention operates on much shorter ones, and the strided memory layout of tabular data makes producing contiguous tensors costly.
Moreover, the hidden dimensions used in current models are small compared to recent language models.
Yet, efficient attention has been studied mostly for one-dimensional sequences, leaving the two-dimensional tabular setting unexplored.
To this end, we create a reproducible benchmarking setup and study the unique characteristics of tabular attention across several backends -- Torch SDPA (efficient and cuDNN), FlashAttention-2/3/4, and the inference-only backends vLLM and SageAttention -- measuring forward and backward throughput across realistic tabular shapes on three GPU generations (A100, H100, B200).
We find that the optimal backend choice differs between column and row attention and varies across hardware as well as model specifics:
While the FlashAttention implementations tailored for each GPU generation perform overall best, they are at times outperformed by cuDNN in the case of column attention at longer sequences with cross-over points depending on the head dimension.
Among inference-only backends, SageAttention performs well for row attention and large sequences beyond 16\,k rows.
Our reproducible benchmark lays the foundation for future improvements to table-native attention.
The self-contained benchmarking and evaluation code is openly available at: \REPOURL
\end{abstract}

\section{Introduction}
\label{sec:introduction}

Tabular data accounts for the majority of data in enterprise machine learning applications~\cite{chui2018notes}.
While gradient-boosted decision trees have long been the dominant paradigm for tabular prediction~\cite{NEURIPS2022_0378c769}, recent pretrained predictive in-context learners such as TabPFN~\cite{tabpfnv2}, TabICL~\cite{tabicl}, Mitra~\cite{mitra}, or ConTextTab~\cite{contexttab} have demonstrated strong, often state-of-the-art performance on public benchmarks~\cite{tabarena}.
At their core, these models are transformer-based and most of them share a common architectural design: alternating \emph{row attention} (attending across data samples) and \emph{column attention} (attending across features), as full cross-cell attention would be prohibitively expensive due to the quadratic scaling of attention with sequence length.
Alternatively, TabICL-like architectures such as TabICLv2~\cite{qu2026tabiclv2} and TabPFN-3~\cite{tabpfn3} perform row- and column-wise attention in a multi-stage approach.

Research on efficient attention, however, has been driven predominantly by the demands of large language models (LLMs).
Implementations such as FlashAttention~\cite{dato2022flashattention, dao2023flashattention2, shah2024flashattention3, zadouri2026flashattention4} target long, one-dimensional sequences -- often with causal masks for autoregressive decoding -- and optimize memory access patterns accordingly.
In parallel, inference-only solutions have emerged for LLM serving: vLLM~\cite{kwon2023efficient} provides Triton-based attention kernels, and SageAttention~\cite{zhang2025sageattention,zhang2025sageattention2,zhang2025sageattention3} offers quantized approximate attention with high forward-pass throughput.
These inference-only backends lack backward-pass support, making them applicable to deployment but not for training.
Tabular data, on the other hand, differs from language data in ways that directly affect attention performance.
First, the sequence lengths for row and column attention are typically asymmetric: the number of columns or features is usually small (tens to low hundreds), while the number of rows can range from hundreds to tens of thousands, limited mostly by GPU memory constraints.
Therefore, both are much smaller than sequences processed in LLMs.
Second, the hidden dimension of current tabular models is relatively small (often in the low hundreds) compared to LLMs, which have hidden dimensions in the thousands, using additional tweaks such as grouped query attention.
Third, tabular attention is bidirectional and non-causal, similar to encoder-only LLMs but unlike current state-of-the-art decoder-only architectures.
And lastly, the standard five-dimensional tabular tensor layout $(B, R, C, H, D)$ -- where $B$ is the batch size, $R$ the number of rows, $C$ the number of columns, $H$ the number of heads, and $D$ the head dimension -- means that column attention operates along a contiguous memory dimension, while row attention requires transposed views with either an explicit \texttt{.contiguous()} copy or a backend that supports strided tensors.

In this work, we provide a systematic characterization and benchmark of attention implementations for tabular data.
Our key contributions are:
(1) We identify the unique characteristics of tabular data and their implications for attention performance.
(2) We benchmark five training backends and two inference-only backends across a range of realistic shapes for row and column attention on three GPU generations (A100, H100, B200).
(3) We analyze the results and provide actionable insights for practitioners building and deploying tabular foundation models.
(4) We release our benchmark code to facilitate future research and development, in particular to encourage attention implementations that are tailored to the unique characteristics of tabular data. To highlight this potential, we perform an AI agent-based optimization of the FA4 kernel implementation on the B200.

\section{Background}
\label{sec:background}

\subsection{General attention}
\label{sec:general-attention}

At its core, the scaled dot-product attention mechanism~\cite{vaswani2017attention} computes, for query, key, and value matrices $\MAT{Q}, \MAT{K}, \MAT{V} \in \mathbb{R}^{N \times D}$,
\begin{equation}\label{eq:attention}
    \OP{Attention}(\MAT{Q}, \MAT{K}, \MAT{V})
    = \OP{softmax}\!\left( \MAT{Q}\MAT{K}\TT / \sqrt{D}\right)\MAT{V} \EQDOT
\end{equation}
It is well known that the naive implementation requires $\bigO{N^2}$ time and memory.
However, several optimized implementations are available that reduce the memory footprint to $\bigO{N}$.
PyTorch's SDPA exposes multiple such backends: the xFormers-based \emph{memory-efficient}~\cite{xFormers2022} and the \emph{cuDNN} backend, which dispatches to NVIDIA's cuDNN library~\cite{chetlur2014cudnn,cudnn}.
Furthermore, the FlashAttention (FA) family of implementations~\cite{dato2022flashattention, dao2023flashattention2, shah2024flashattention3, zadouri2026flashattention4} provides highly optimized attention kernels for NVIDIA GPUs.
Whereas FA2 is optimized for Ampere GPUs (A100), FA3 targets Hopper GPUs (H100) and the recent FA4 implementation is primarily tuned for the Blackwell architecture (B200).
Beyond the FlashAttention family, inference-only alternatives have emerged:
SageAttention~\cite{zhang2025sageattention, zhang2025sageattention2, zhang2025sageattention3} provides quantized approximate attention with high forward-pass throughput, and the vLLM project~\cite{kwon2023efficient} includes Triton-based attention implementations optimized for LLM serving.
These inference-only backends support only forward passes, making them relevant for deployment but not for training.
We provide more details on the individual backends in \cref{sec:attention-backends}.

\subsection{Tabular attention}
\label{sec:tabular-attention}

There are multiple ways to represent tabular embeddings for transformer-based models.
With a batch of $B$ tables, with $R$ rows (samples) and $C$ columns (features), and a hidden dimension of $d{=}H{\cdot}D$ using $H$ heads of dimension $D$, the standard layout is a five-dimensional tensor of shape $(B, R, C, H, D)$, which we refer to as the \emph{row-first} layout.
Similar to attention used in Vision Transformers~\cite{dosovitskiy2020image}, which also operate on 2D sequences as opposed to 1D sequences encountered in language modeling, applying full attention across both the row and the column dimension is infeasible, due to its quadratic scaling.
However, approaches used in Vision Transformers, such as patching or hierarchical attention~\cite{vaswani2017attention}, do not natively translate to the tabular domain, due to its lack of locality.
That is, most tabular applications are considered to be invariant or equivariant against row and column order permutations.
Therefore, in most recent models, attention is applied alternately along the row and column dimensions instead:
\emph{Column attention} attends across features for each sample, and \emph{row attention} attends across samples for each feature.
Hence, for each attention operation, the sequence length is either $C$ or $R$, and the effective batch size is $B \cdot R$ or $B \cdot C$, respectively.
The tensor has to be reshaped accordingly to apply the attention operation along the correct dimension, depending on the backend used.
The PyTorch-native backends require an input shape of $(\cdots, H, L, D)$ with sequence length $L$, requiring non-contiguous reshapes for both row and column attention.
That is, for column attention, the tensor is reshaped to $(B \cdot R, H, C, D)$, which can be represented as a zero-copy view, only requiring a single copy-inducing \texttt{.contiguous()} call for the output.
The row attention, on the other hand, requires a reshape to $(B \cdot C, H, R, D)$, which cannot be performed using a zero-copy view, needing a \texttt{.contiguous()} call for each of the queries, keys, and values.
Depending on the size of the tensor, this can lead to significant overhead, as we will observe in the results.
This attention pattern holds for all TabPFNv2-like models but not directly for TabICL which uses a dedicated two-stage attention approach.
Other backends, notably the FA family, use a more compact stride layout with innermost $H$ and $D$ axes.
That is, for column attention the tensor can natively be reshaped to $(B \cdot R, C, H, D)$ which is a zero-copy view of the original tensor.
In addition, even the row attention reshape $(B \cdot C, R, H, D)$ can be performed as a zero-copy transpose view and does not require an explicit copy.
Hence, it only involves one final \texttt{.contiguous()} call to restore a contiguous layout after one iteration of row and column attention for subsequent operations -- much fewer memory operations than required by the SDPA backends.

In either case, an asymmetry between column and row attention exists:
Column attention operates along a contiguous memory dimension with shorter sequences, while row attention operates along a non-contiguous dimension with longer sequences.
The same argument holds vice versa if using a column-first representation of shape $(B, C, R, d)$, which would make row attention contiguous and column attention non-contiguous.
While the overhead for the memory copy is identical, the row-first layout might still be preferred:
As the number of rows is typically much larger than the number of columns, the stride is more compact in the row-first layout.
This makes L1 cache hits generally more likely and thus can lead to better performance in other layers of the model.
Hence, in the following, we will focus on the row-first layout, which is also the standard layout used in many current tabular models.
We analyse this in more detail in \cref{sec:layout-ablation}.

Finally, note that the hidden dimension of current tabular models is much smaller than that of LLMs. This can affect the performance of different backends, and some backends impose restrictions on the number of dimensions they can operate on.
For example, FA4 only supports head dimensions of \num{128} for Hopper GPUs and \num{64} or \num{128} for Blackwell GPUs, which is partially beyond what is typically used in tabular models:
TabPFN uses six heads with \num{32} dimensions, whereas TabICL uses eight heads at \num{16}, and ConTextTab uses \num{12} heads at \num{64} head dimension.

\section{Benchmark and Experiments}
\label{sec:experiments}

To investigate the influence of the previously discussed table-native characteristics on attention as used in recent tabular foundation models, we constructed a self-contained reproducible tabular attention benchmark, measuring latency and memory consumption for isolated kernel throughput on synthetic tabular tensors as well as end-to-end validations.

\custompar{Tensor shapes}
We benchmark both column and row attention using the row-first tensor layout $(B, R, C, H, D)$.
We focus on the case with batch size $B{=}\num{1}$, which allows us to benchmark a wide range of realistic shapes without being limited by GPU memory, but investigate batch size sweeps in \cref{sec:additional-results}.
Note that we nevertheless use the SDPA implementation with \texttt{.contiguous()} calls for the row attention even though it is not required for $B=\num{1}$, to have a realistic comparison.
For \emph{column attention}, we fix $R{=}\num{1024}$ rows and vary the number of columns $C{\in}\{\num{16}, \num{32}, \dots, \num{2048}\}$, yielding an effective batch size of $B{\CDOT}R {=} \num{1024}$ and sequence length equal to $C$.
For \emph{row attention}, we fix $C{=}\num{64}$ columns and vary $R{\in}\{\num{32}, \num{64}, \dots, 131072\}$, giving an effective batch size of $B \CDOT C{=}\num{64}$ and sequence length equal to $R$.
These ranges reflect realistic tabular datasets, where feature counts are typically moderate and sample counts vary widely.
For example, across the TabArena, CARTE, and TALENT benchmarks~\cite{tabarena, CARTE, talent}, we find tables ranging from roughly 4 to 2\,k columns and 500 to 100\,k rows.
For the main discussion, we focus on $H{=}\num{12}$ heads with head dimension $D{=}\num{64}$, a configuration used in ConTextTab~\cite{contexttab}, but also provide results for other configurations -- ranging from hidden dimensions \num{128} to \num{2048} -- to include settings used in other models in \cref{sec:additional-results}.

\custompar{Backends}
We evaluate five backends that support both forward and backward passes:
(1)~{SDPA (efficient)}, PyTorch's xFormers-based backend;
(2)~{SDPA (cuDNN)}, using NVIDIA's cuDNN library;
(3)~{FlashAttention-2}, optimized for Ampere GPUs;
(4)~{FlashAttention-3}, which supports strided tensors and targets Hopper GPUs; and
(5)~the recent {FlashAttention-4}, tailored to Blackwell GPUs but imposing a minimum sequence length of 128 for Hopper GPUs.
In addition, we evaluate two inference-only backends for forward-pass throughput:
(6)~{vLLM}, using Triton-based attention kernels; and
(7)~{SageAttention}, providing quantized approximate attention.

All implemented backend wrappers for row and column attention are tested against the naive \texttt{math} PyTorch backend to verify correct implementation behaviour, and benchmarked for both the forward and backward pass, while inference-only backends are benchmarked for the forward pass only.

\custompar{Attention patterns}
As previously mentioned, we focus on the case of 2D-interleaved attention as used by TabPFNv2, Mitra, and ConTextTab.
We do so in a slightly idealized fashion by ignoring the train-test (context-query) split that is handled differently among model families:
while TabPFN and Mitra use a cross-attention formulation, where the row attention Q corresponds to samples from both train and test and KV comes from train alone, ConTextTab uses a masking approach.
Instead, we evaluate standard full bi-directional attention.
As such our results can be considered representative of the zero-query limit and close to the single-query inference case.

\custompar{Measurement protocol}
The measurement protocol closely follows the implementation in the FlashAttention repository\footnote{\URL{github.com/Dao-AILab/flash-attention}}.
Timing uses CUDA events recorded around each kernel call, providing sub-microsecond GPU-side timing that is unaffected by CPU-side overhead.
Before each timed repetition, a dedicated GPU buffer is written to flush the L2 cache, ensuring that each measurement reflects a cold-cache kernel execution rather than cached memory access patterns.
The GPU clock frequency is stabilized prior to any measurement by running a sequence of large matrix multiplications.
Each configuration is then measured over multiple repetitions, ranging from three to 50 depending on the sequence length, preceded by five warmup iterations.
The backward pass time is isolated by subtracting the mean forward time from a combined forward+backward measurement, with the variance computed by summing the individual variances of the two independent runs, analogous to the timings in FlashAttention.
We report throughput in TFLOPS, computed as $4 \CDOT B_{\text{eff}} \CDOT H \CDOT L \CDOT L \CDOT D \,/\, t$ for the forward pass and $\num{2.5}\TIMES$ the forward FLOPs for the backward pass following the convention in the FlashAttention benchmark.
Throughout, we report the mean and $\pm 1\sigma$ across repetitions.
In addition, we measure peak memory consumption.

\custompar{Hardware and software}
We evaluate the benchmark on the three most recent NVIDIA GPU generations: an NVIDIA A100 (\SI{80}{\giga\byte} HBM2e), an NVIDIA H100 (NVL, \SI{80}{\giga\byte} HBM3), and an NVIDIA B200 (NVL, \SI{192}{\giga\byte} HBM3e).
Not all backends are available on all GPUs: FA3 and FA4 require Hopper or newer, and FA4 is primarily optimized for Blackwell, requiring a head dimension of 128 on H100 or at least 64 on B200.
Notably, this makes FA4 incompatible with recent models such as TabPFNv2 or TabICL, due to their small used head dimensions, regardless of the GPU architecture used, whereas TabPFN 2.5 and 2.6 and ConTextTab use a head dimension of 64.
Moreover, SageAttention2 is only available on Hopper, while SageAttention3 does support Blackwell GPUs, but only the consumer-grade RTX versions, which we do not include here.

Throughout, we use the latest supported PyTorch and CUDA versions for each backend: PyTorch 2.10 with CUDA 13.0 and cuDNN 9.15.1 for SDPA, FA3, and FA4, falling back to PyTorch 2.8 with CUDA 12.8 for FA2.
For all backends, we use the kernels precompiled for the corresponding PyTorch and CUDA version or compile them from scratch if required.
All tensors use \texttt{bfloat16} precision, and attention is non-causal throughout as is common for tabular foundation models.

\custompar{Benchmark infrastructure}
Each backend is run using a custom environment and within a single process.
The setup is easy to use, with predefined\;\texttt{make}\;hooks and per-backend dependency definitions using \texttt{uv}, ensuring a consistent backend-specific setup for each benchmarking round.
Results are flushed to disk as JSON after every completed shape, so a mid-run failure loses at most the single shape currently in flight.
Each result file encodes the full configuration (GPU, PyTorch and CUDA versions, dtype, head configuration, sweep parameters, repetition count) alongside the per-shape timing statistics, making results self-contained and reproducible.
The benchmark code, all result files, and plotting scripts are released at: \REPOURLSHORT.

\section{Results}
\label{sec:results}

\begin{figure}
    \centering
    \begin{subfigure}[b]{0.97\textwidth}
        \centering
        \includegraphics[width=\textwidth]{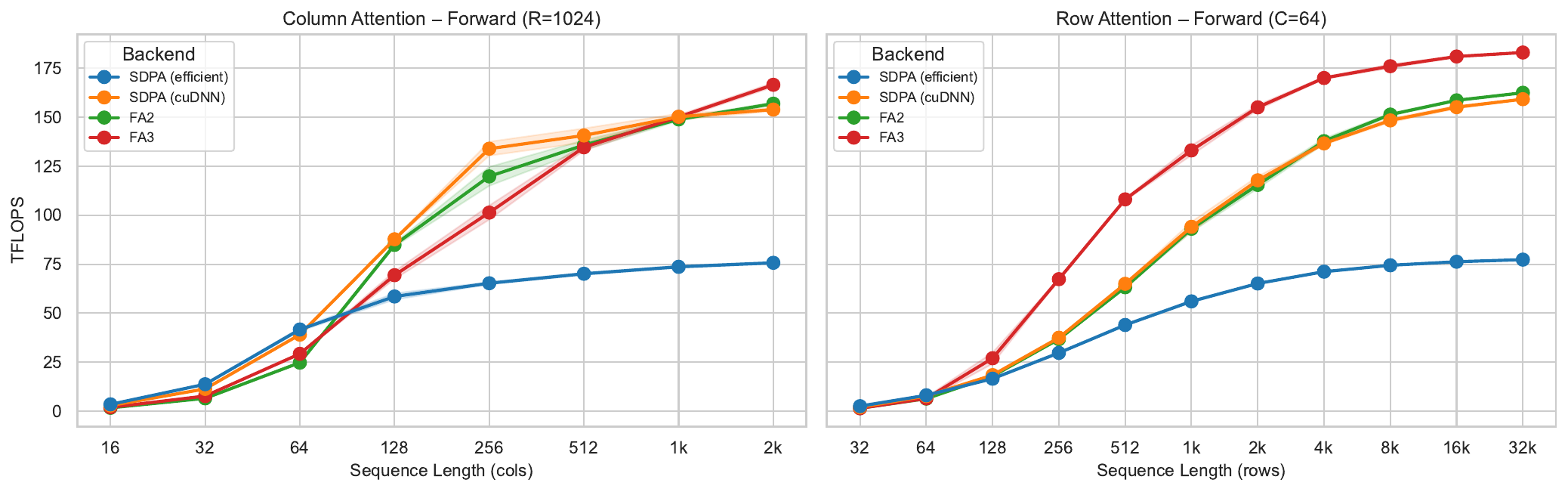}
        \caption{NVIDIA A100 (PCIe).}
        \label{fig:comparison-a100}
    \end{subfigure}
    \begin{subfigure}[b]{0.97\textwidth}
        \centering
        \includegraphics[width=\textwidth]{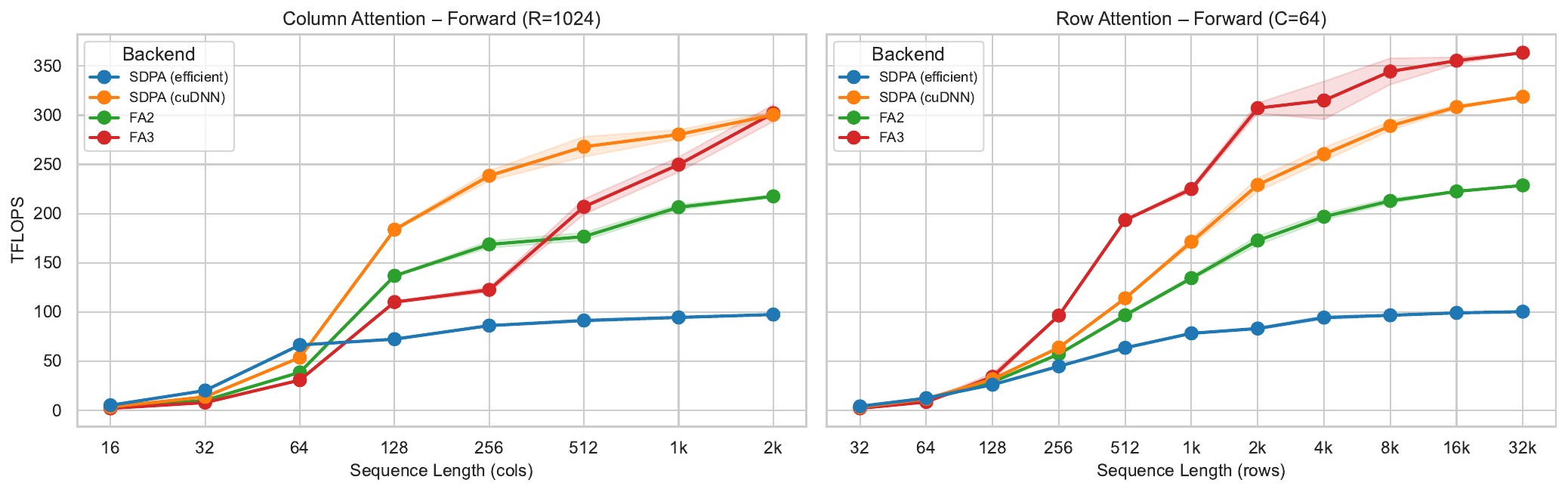}
        \caption{NVIDIA H100 (NVL).}
        \label{fig:comparison-h100}
    \end{subfigure}
    \begin{subfigure}[b]{0.97\textwidth}
        \centering
        \includegraphics[width=\textwidth]{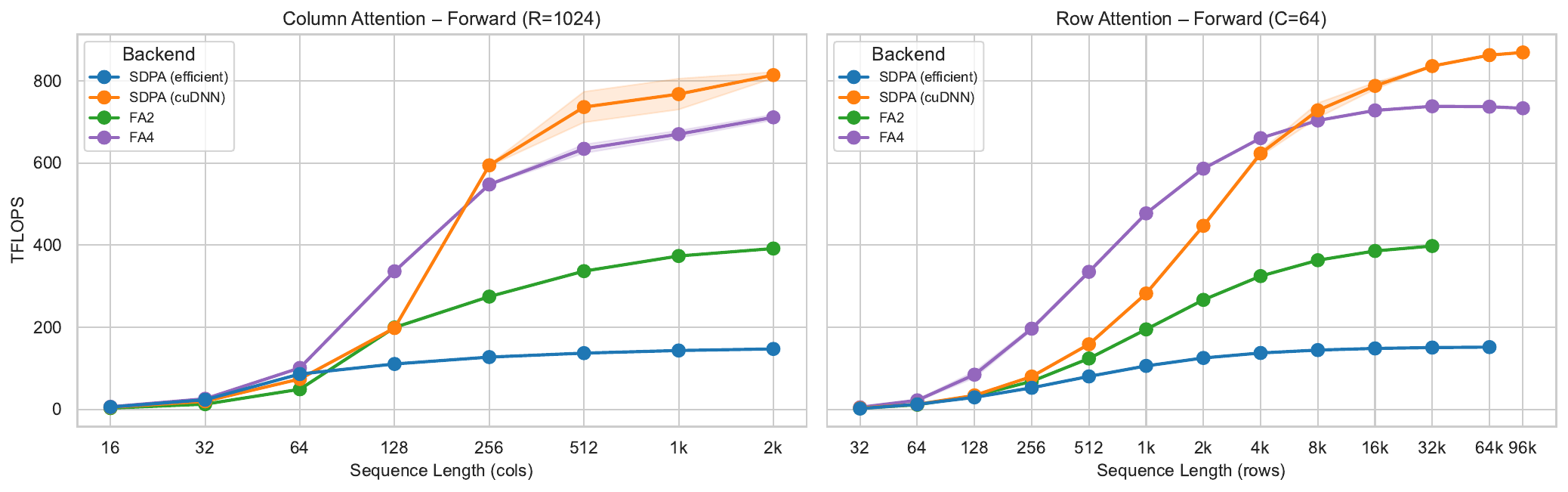}
        \caption{NVIDIA B200 (NVL).}
        \label{fig:comparison-b200}
    \end{subfigure}
    \caption{Forward throughput for column attention (left) and row attention (right) across three GPU generations.
    Configuration: $H{=}12$ heads, $D{=}64$, \texttt{bfloat16}.
    Note that FA3 is unavailable on B200 and FA4 does not support $D{=}64$ on H100. Missing measurements due to GPU memory constraints.}
    \label{fig:gpu-comparison}
\end{figure}

\cref{fig:gpu-comparison} shows the forward throughput of all investigated backends for both column and row attention across three GPU generations in the case of 12 heads with 64 head dimension.
We report full results, including backward pass throughput, peak memory, and other shape configurations in \cref{sec:additional-results}.

\custompar{Column attention}
While details depend on the GPU used, some general patterns can be observed:
As only a single \texttt{.contiguous()} call is required and the attention operates on the native view, we mostly observe the expected native backend scaling with the sequence length.
SDPA (efficient) dominates the very small sequence lengths, likely due to a minimal kernel launch overhead, whereas SDPA (cuDNN) achieves very strong throughput across all investigated GPU families.
Between $C{=}\num{16}$ and \num{64}, SDPA (efficient) is up to $2\TIMES$ faster than the remaining backends on all investigated GPUs, however in absolute throughput differences, the gap is comparably small.

For the A100, cuDNN is only slightly surpassed by FA3 for very long sequence lengths.
Notably, FA2, being optimized for Ampere, shows generally slightly worse performance than the more recent cuDNN backend, but the differences between FA2, FA3 and cuDNN are mostly negligible here.

On the H100 GPU, cuDNN again dominates the forward-pass column attention performance for sequence lengths between 128 and 1\,k, being outperformed slightly by FA3 for longer sequences.
Surprisingly, however, FA3 performs slightly worse than FA2 in the practically important range of up to 256 columns.
Generally, on the H100, the cuDNN backend shows the overall best column attention throughput, with significant speedups over FA3 in the 128--512 columns region.
Note that FA4 is missing in this setup as it does not support a head dimension of 64 on the H100.

For the Blackwell B200, the recent FA4 backend dominates the range up to 256 columns, being significantly outperformed, however, for longer sequences by the cuDNN backend.
Overall, the throughput on B200 is impressive: with cuDNN and FA4 backends achieving more than $2\TIMES$ throughput as compared to the best backend on H100 and more than $5\TIMES$ as compared to A100.

While the details differ between the GPU families, cuDNN performs either state-of-the-art or close to it across all sequence lengths and GPU architectures investigated, making it a good cross-platform choice, with some improvements made by FA4 in the case of Blackwell GPUs for sequences below 256 columns.
As discussed in \cref{sec:additional-results}, similar observations hold for the backward pass.

\custompar{Row attention}
The picture changes substantially for row attention:
On the A100 and H100 GPUs, FA3 dominates across all sequence lengths except for the very short ones below $R{=}128$, achieving up to $3.5\TIMES$ speedup over the SDPA baseline and around $1.5\TIMES$ speedup over cuDNN.
The key driver of this advantage is more compact stride support, avoiding costly copies for the queries, keys, and values, as well as better performance of the tuned kernel with longer sequences which in particular the FA implementations are optimized for.

On the B200, the recent FA4 backend dominates the evaluated range but is surprisingly outperformed by cuDNN on very long sequence lengths beyond 8\,k rows.
In the regime between $R{=}128$ and $\num{4096}$, which is currently the most relevant for training tabular models, this copy overhead seems to have a significant impact, with diminishing importance as the sequence length increases.
Since the memory copy is $\bigO{N}$ in time, while the attention kernel is $\bigO{N^2}$, the relative overhead of the copy decreases as the sequence length increases, which is reflected in the diminishing relative gap between FA3/FA4 and cuDNN for longer sequences.
Note that this crossover behaviour depends on the batch size and hidden dimension, determining the total tensor size and thus the copy time, as we will later discuss.
In the backward pass, the behaviour is similar.

Overall, in the case of row attention, optimized backends such as FA3 and FA4 show substantial gains over the cuDNN baseline, unlike what we have observed for the (more conventional) case of column attention, highlighting the unique challenges and characteristics of applying attention on 2D tabular embedding sequences.
Further results, using different head dimensions and head counts as well as speedup plots, are provided in the appendix, showing consistent patterns with the main results.

\begin{figure}
    \centering
    \includegraphics[width=0.97\textwidth]{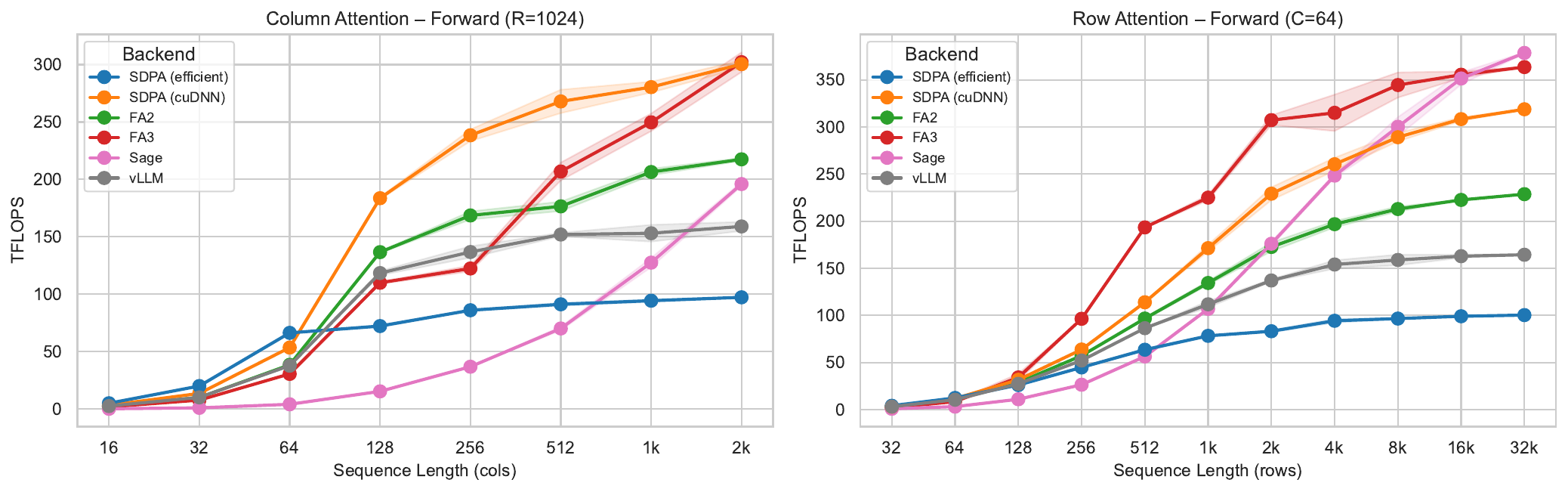}
    \caption{Forward throughput including inference-only backends (vLLM, SageAttention) on the H100.
    Configuration: $H{=}12$ heads, $D{=}64$, \texttt{bfloat16}.
    Error bands show $\pm 1\sigma$ over \num{50} repetitions.}
    \label{fig:inference-only}
\end{figure}

\custompar{Inference-only backends}
In addition to the general-purpose backends evaluated thus far, we show results for two inference-only backends that lack backward pass support in \cref{fig:inference-only}: vLLM's Triton-based prefill kernel and SageAttention~2++.
Here, the results are only reported for the H100 as SageAttention backends are not available on B200 as previously noted.

On the H100, SageAttention achieves sub-par throughput for column attention for practically all important ranges below 2\,k columns, potentially due to a large kernel launch overhead.
However, an impressive throughput scaling can be observed.
This becomes dominant for row attention, due to the larger sequence lengths: beyond 16\,k rows, SageAttention overtakes the other backends, scaling efficiently up to 128\,k rows in the case of smaller head dimensions.
The vLLM kernel, while designed for LLM prefill workloads, shows reasonable performance for column attention but does not match the specialized backends for row attention.
These results suggest that using SageAttention for deployed inference models can achieve throughput gains when applied for row attention at long sequence lengths, e.g.\ combined with a cuDNN backend for column attention and for smaller sequences, in the case of the H100 GPU.
However, note that SageAttention uses an approximate attention implementation.
Therefore, the effect on the model's predictive performance needs to be evaluated for the model at hand, which we investigate in the Appendix for TabPFNv2 and ConTextTab.

\begin{figure}
    \centering
    \includegraphics[width=0.97\textwidth]{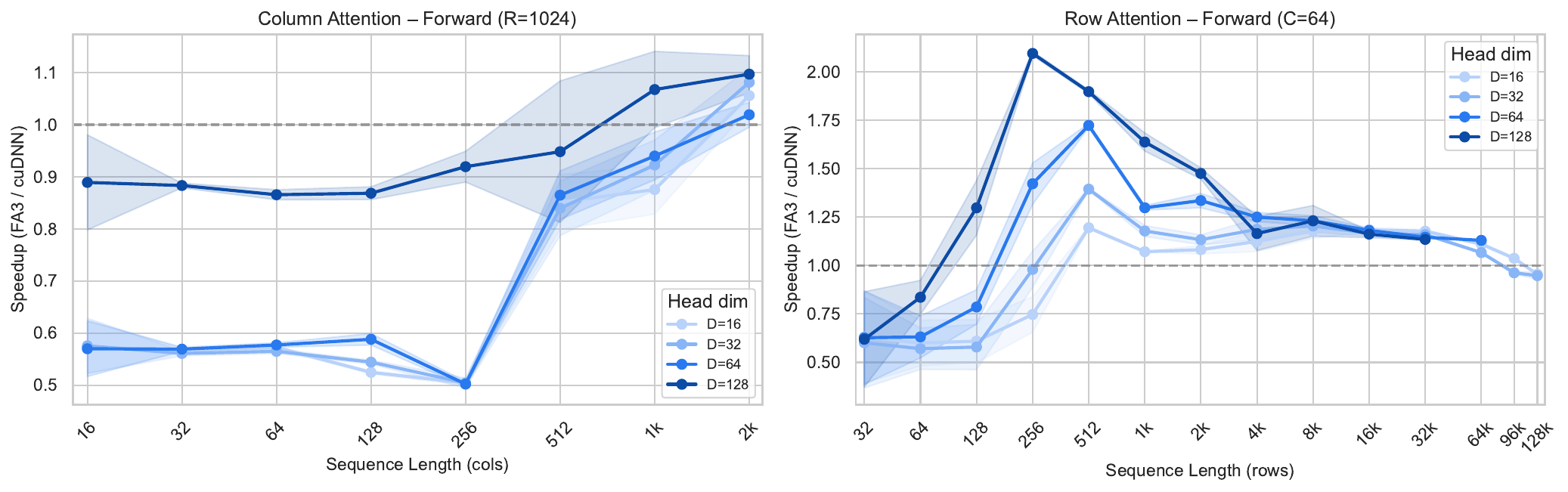}
    \caption{Speedup of FA3 over SDPA (cuDNN) for different head dimensions on the H100.
    Configuration: $H{=}8$ heads, $D \in \{16,\dots, 256\}$, \texttt{bfloat16}.
    Darker shades indicate larger head dimensions.
    Values above the dashed line indicate FA3 is faster.
    The relative advantage of FA3 in row attention increases with the head dimension, as the \texttt{.contiguous()} copy cost scales with $D$.}
    \label{fig:headdim-ablation}
\end{figure}

\custompar{Head dimension ablation}
Finally, as noted before, current tabular foundation models have widely ranging hidden dimension and head count details.
While TabPFN operates on six heads with \num{32} dimensions, TabICL uses \num{8} heads at \num{16}, and ConTextTab uses \num{12} heads at \num{64} head dimension.
We have so far discussed the case with 12 heads and 64 head dimension, corresponding to a hidden dimension of 768.
To ablate its impact, we show the speedup of FA3 over SDPA (cuDNN) across different head dimensions with a fixed setup of 8 heads evaluated on the H100 GPU.
The results are shown in \cref{fig:headdim-ablation}.
For column attention, FA3 is generally slower than cuDNN for practically relevant head dimensions and sequence lengths.
For much larger head dimensions, however, FA3 may achieve up to $1.4\TIMES$ speedup over cuDNN, whereas the performance seems to converge to similar throughputs for longer sequence lengths.

For row attention, we see a clear pattern:
The advantage of FA3 grows with the head dimension.
Whereas cuDNN performs better for smaller sequence lengths, this cross-over point shifts monotonically to smaller sequence lengths with larger head dimensions.
This likely correlates with the increased copy overhead:
The three \texttt{.contiguous()} copies avoided by FA3's strided tensor support scale linearly with $D$, making the overhead more pronounced at larger head dimensions.
For very large head dimensions, this can result in up to $2.4\TIMES$ speedup with diminishing returns as the sequence length grows and the quadratic scaling of the attention calculation dominates over the linear scaling of the copy overhead.
Hence, for models using smaller head dimensions, the gain of FA3 over cuDNN may be limited.
However, as the development of foundation models continues, larger dimensions may be used which benefit more from optimized FA implementations.

\custompar{Practical recommendations}
As we have seen, there is no one-fits-all solution. The unique characteristics of tabular attention and strong dependence on model specifics (head dimension), GPU family, as well as intended use range (number of rows and columns) influence the optimal backend choice, with differences between cross-row and cross-column attention.
Based on these results, we generally recommend a mixed strategy for tabular attention in the presented setting (H100 GPU, $H{=}12$, $D{=}64$, non-causal attention with \texttt{bfloat16} precision):
Using the cuDNN backend for \textit{column attention} and FlashAttention-3 for \textit{row attention} across all sequence lengths, potentially falling back to SDPA (efficient) for very short sequence lengths.
For inference-only deployments, approximate implementations such as SageAttention may greatly speed up row attention in particular, however, its influence on the model's predictive performance needs to be carefully examined.
All considerations depend on the model at hand, namely its choice of the head count and head dimension, and our benchmark helps practitioners find optimal backends for their specific needs.
Similar conclusions may be drawn from the additional results provided across a wide range of model configurations and GPUs, tailored to individual specific needs.

\begin{figure}
    \centering
    \includegraphics[trim=0 0 0 10mm, clip, width=0.97\textwidth]{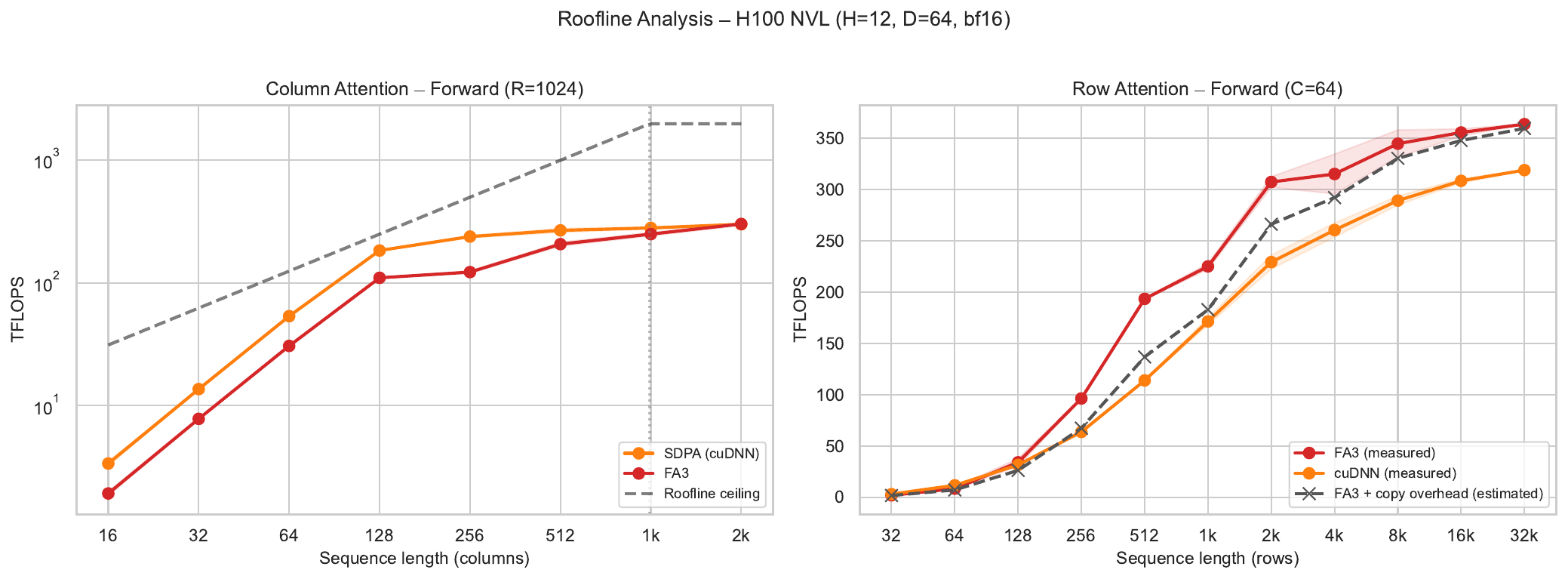}
    \caption{Roofline analysis on the H100.
    \emph{Left:} Column attention throughput vs.\ the memory-bound roofline ceiling.
    \emph{Right:} Row attention showing FA3 (no copies), cuDNN (with copies), and the estimated throughput of FA3 if degraded by a copy overhead. Configuration: $H{=}12$, $D{=}64$.}
    \label{fig:roofline-analysis}
\end{figure}

\section{Analysis}
\label{sec:analysis}

To provide a quantitative framework for the observed differences, we analyze them via the roofline model and quantify the \texttt{.contiguous()} copy overhead.
In the following, we focus on the H100 and FA3 case, as the most established and practically relevant at the moment.

\custompar{Operational intensity of tabular attention}
The operational intensity (OI) of the forward attention pass -- defined as FLOPs per byte of memory traffic -- determines whether a kernel is memory-bound or compute-bound.
For non-causal attention with effective batch size $B_{\text{eff}}$, $H$ heads, sequence length $N$, and head dimension $D$, the forward FLOPs are $4 \CDOT B_{\text{eff}} \CDOT H \CDOT N^2 \CDOT D$.
The minimum HBM transfer (reading Q, K, V and writing O in the IO-optimal case) is $4 \CDOT B_{\text{eff}} \CDOT H \CDOT N \CDOT D \CDOT 2$ bytes in \texttt{bfloat16}.
Their ratio yields:
\begin{equation}\label{eq:oi}
    \text{OI} = \frac{4 \CDOT B_{\text{eff}} \CDOT H \CDOT N^2 \CDOT D}{4 \CDOT B_{\text{eff}} \CDOT H \CDOT N \CDOT D \CDOT 2} = \frac{N}{2} \EQDOT
\end{equation}
The ridge point -- where the memory-bound slope meets the compute-bound ceiling -- is determined by the ratio of peak compute to peak bandwidth.
For the H100 NVL (ca.\ \SI{2000}{\tera\text{FLOPS}} BF16 tensor core, \SI{3.9}{\tera\byte\per\second} HBM3), the ridge point lies at approximately \num{500}~FLOP/byte, corresponding to a sequence length of $N \approx \num{1000}$.
This immediately reflects the previously observed patterns:
Column attention, with typical sequence lengths of $C{=}\num{16}$--$\num{256}$, operates at $\text{OI}{=}\num{8}$--$\num{128}$, placing it firmly in the \emph{memory-bound} regime.
Row attention at longer sequences ($R{=}\num{512}$--$\num{16384}$, $\text{OI}{=}\num{256}$--$\num{8192}$) transitions toward the compute-bound regime.
\cref{fig:roofline-analysis} (left) confirms this: for column attention, the roofline ceiling (dashed line) provides an upper bound on achievable throughput, with cuDNN achieving performance close to the memory-bound ceiling at $C{=}\num{128}$.
The gap between measured throughput and the theoretical ceiling at longer column sequences reflects insufficient parallelism: although $B_{\text{eff}}{=}\num{1024}$, the per-head problem size at larger $C$ remains too small to fully saturate the memory subsystem.
Larger effective batch sizes would shift this saturation point to longer sequences.

\custompar{Quantifying the copy overhead for row attention}
For row attention, some backends require contiguous inputs (SDPA efficient and cuDNN) and must perform three \texttt{.contiguous()} copies for Q, K, and V.
The total copy time for tensors of shape $(B_{\text{eff}}, H, N, D)$ in \texttt{bfloat16} is:
\begin{equation}\label{eq:copy-time}
    t_{\text{copy}} = \frac{3 \CDOT B_{\text{eff}} \CDOT H \CDOT N \CDOT D \CDOT \SI{2}{byte}}{\beta_{\text{copy}}(N)} \EQCOMMA
\end{equation}
where $\beta_{\text{copy}}(N)$ is the achieved bandwidth (in byte/s) for strided-to-contiguous copies, which we measure at each sequence length using the exact tensor shapes from the benchmark.
Crucially, $\beta_{\text{copy}}$ depends on tensor size: at the smallest shapes ($N{=}\num{32}$, \SI{9}{\mega\byte}), we measure only $\beta_{\text{copy}} \approx \SI{783}{\giga\byte\per\second}$, while larger tensors ($N{\geq}\num{2048}$, ${\geq}\SI{576}{\mega\byte}$) saturate at $\beta_{\text{copy}} \approx \SI{1.45}{\tera\byte\per\second}$.
In either case, $\beta_{\text{copy}}$ remains well below the peak HBM bandwidth of \SI{3.9}{\tera\byte\per\second} because the non-sequential read pattern required for the contiguous copy of the strided tensors prevents full bandwidth utilization.

\cref{fig:roofline-analysis} (right) shows the row attention throughput for FA3 (which avoids copies via strided tensor support), cuDNN (which requires copies), and an estimated line computed as:
$\text{TFLOPS}_{\text{estimated}} = \text{FLOPs} \,/\, (t_{\text{FA3}} + t_{\text{copy}})$,
\ie the throughput that FA3's kernel would achieve if additionally burdened by the measured copy overhead.
This estimated line tracks the cuDNN measurements closely at medium sequence lengths ($N{=}\num{256}$--$\num{1024}$), where the copy overhead explains the majority of the throughput gap.
At longer sequences ($N{>}\num{2048}$), the estimated line exceeds the measured cuDNN throughput, indicating that FA3's kernel is additionally more efficient for longer sequences -- likely due to better tiling and work distribution optimized for this regime.
Since the copy time scales as $\bigO{N}$ while attention computation scales as $\bigO{N^2}$, the relative copy overhead diminishes at longer sequences, with the remaining gap reflecting FA3's superior kernel efficiency in this regime.
Conversely, larger head dimensions $D$ increase the copy overhead linearly, explaining the pattern observed in the head-dimension ablation (\cref{fig:headdim-ablation}): the advantage of FA3 over cuDNN grows with $D$ because the copy tax increases proportionally.

\custompar{Implications}
This analysis provides a principled justification for the mixed-backend strategy:
(1) Column attention is memory-bound and involves no input copies in the row-first layout.
The cuDNN backend, being highly optimized for memory-bound workloads with low kernel launch overhead, naturally dominates. (2) Row attention benefits from strided tensor support (avoiding runtime overhead from copies) \emph{and} from FA3's superior kernel design for longer sequences.
Moreover, the analysis predicts that on future hardware with higher compute-to-bandwidth ratios (higher ridge points), the memory-bound regime will extend to longer column sequences, while the relative copy overhead for row attention will grow -- further favoring backends with strided tensor support.

\section{End-to-end Validation}
\label{sec:e2e-validation}

To illustrate that isolated kernel speedups translate to end-to-end speedups, we investigate the full inference timing of TabPFN-v2.6 under different attention backends using the H100 GPU.
We note that this end-to-end evaluation is meant as an initial, illustrative validation rather than being comprehensive: it is inference-only and restricted to a single model and GPU as it requires individual model surgery that is prone to errors or omitting architecture-specific details.
We use the publicly available model code\footnote{\url{https://github.com/PriorLabs/TabPFN/tree/v7.1.1}} and patched the attention layer to utilize our attention wrappers as well as comparing it to the original dispatch path.
We measure the end-to-end timings of a full `predict` path using the same synthetic data as before, sweeping sample and feature counts.
We focus on the setting using a single query row but note that batched-query results are conceptually very similar.

\begin{figure}
    \centering
    \includegraphics[width=0.48\textwidth,trim={177mm 130mm 0 8mm},clip]{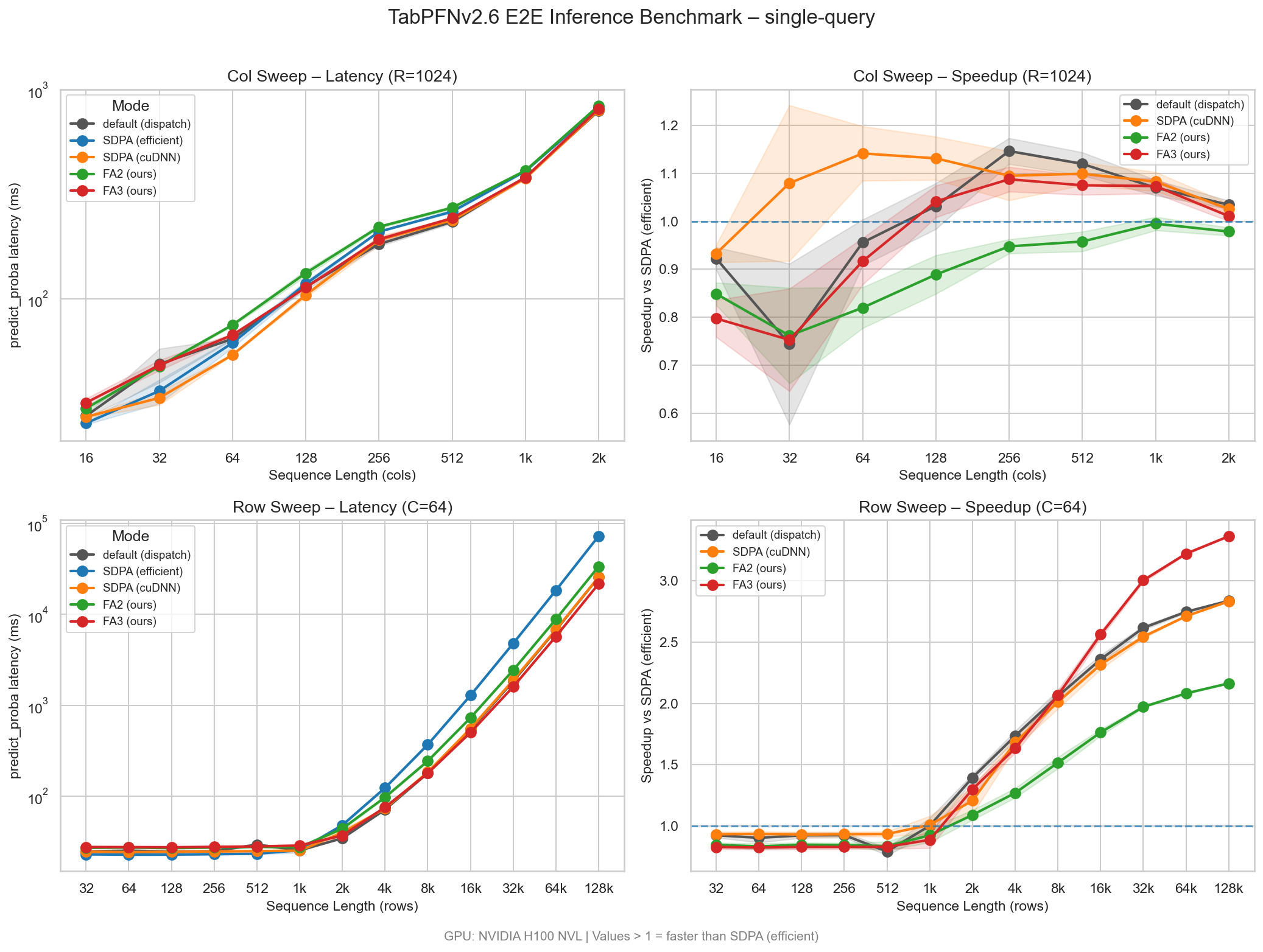}
    \hspace{1mm}
    \includegraphics[width=0.48\textwidth,trim={177mm 10mm 0 137mm},clip]{e2e/e2e_single-query_NVIDIA_H100_NVL}
    \caption{End-to-end speedup measurements of a full inference call for TabPFN-v2.6 across different investigated backends including TabPFN's default dispatch using an H100 GPU.}
    \label{fig:speedup-e2e}
\end{figure}

The results are depicted in \cref{fig:speedup-e2e}.
We see that isolated kernel speedups generally transfer to end-to-end inference speedups.
We observe speedups of up to 15\% at a fixed sample size across different column lengths, with cuDNN performing best in the narrow-table regime.
At longer sequences, cuDNN, FA3 and the default dispatch of TabPFN tend to perform similarly.
For the row sweep, we observe strong speedups vs the naive SDPA baselines with speedups of more than 300\%, and FA3 showing significant gains at longer sequences as compared to TabPFN's default dispatch.
Overall, the results provide initial evidence that differences in the isolated kernel performance translate to end-to-end inference timings and that the default dispatch of TabPFN-v2.6 is not optimal in the narrow- as well as very-long-table regime.
We emphasize that this is an inference-only evaluation on a single model and GPU, and thus serves as supportive evidence for the practical relevance of our kernel benchmark rather than a comprehensive end-to-end validation across architectures.

Finally, we use these insights to define a ``combined'' backend, using cuDNN for cross-column attention, the default memory efficient SDPA for cross-row attention for less than 2\,k rows, and FA3 for row-attention with larger sample sizes.
We validate this combination end-to-end and measure speedups across all tasks of the TabArena benchmark (using its single-fold ``lite'' variants).
We observe a mean speedup of 7.4\% over TabPFN's default dispatch, with speedups ranging to up to 30\%, corresponding to reduced latencies of up to 1\,s.
Due to the different table sizes within the benchmark, speedups vary wildly, and reporting mean/median aggregates often does not highlight nuanced scaling behavior.
We report further detailed results in \cref{sec:additional-results}.

\section{Conclusion}
\label{sec:conclusion}

We have presented a systematic benchmark of attention implementations for tabular foundation models, covering five training backends and two inference-only backends across realistic shapes on three GPU generations (A100, H100, B200) with \texttt{bfloat16} precision.
We have discussed the unique characteristics of tabular attention as compared to attention in language models or Vision Transformers, including the asymmetry between row and column attention and the impact of tensor contiguity and head dimension on attention throughput.
Our results show that the optimal backend choice depends critically on the attention pattern: SDPA (efficient) excels at the short column-wise sequences, cuDNN dominates on longer column sequences, while FlashAttention-3/4 dominates row attention by avoiding costly tensor copies and showing better sequence scaling, depending on the GPU family at hand.
The performance gap between the best and worst backend can exceed $3\TIMES$ for a given shape, underscoring the importance of backend selection in tabular deep learning.

A key insight is that the contiguity overhead from \texttt{.contiguous()} calls -- often dismissed as an implementation detail -- has a measurable impact on attention throughput as it dominates the runtime in the medium sequence length regime, depending on the batch size and hidden dimension.
Backends that support more compact strides for their input, such as FA3 and FA4, avoid this overhead to a large extent, providing a consistent advantage.
We hope that these findings encourage research and development of custom attention strategies tailored to the unique characteristics of tabular data, such as optimized kernels for the typical shapes and patterns of row and column attention, or novel approaches that can further mitigate the contiguity overhead.
To highlight one future application, we perform a preliminary AI Agent-based optimization of the FlashAttention-4 kernel on the B200 (see \cref{sec:agent-optimization}), showing promising initial results, demonstrating that our benchmark enables a closed-loop workflow specifically for optimizing attention for tabular embeddings.

With our self-contained benchmark and evaluation code we hope to spur research for the development of table-adapted kernels in the future, being the core of current tabular foundation model architectures.

\custompar{Limitations}
Our benchmark focuses primarily on isolated attention kernels with synthetic tensors, complemented by an illustrative end-to-end validation on a single model and GPU; it thus does not fully capture interactions with surrounding layers or system-level effects such as multi-GPU or cross-node communication.
We evaluate only \texttt{bfloat16} precision and non-causal attention; other precisions or masked variants may exhibit different relative performance, however these are currently of less practical relevance in the case of tabular models.
We further focus on the interleaved 2D attention shared by TabPFN-v2-, Mitra-, and ConTextTab-like models, evaluating standard bidirectional attention rather than the cross-attention or masking formulations these architectures use in detail, and do not cover the multi-stage row- and column-wise attention of TabICL-like architectures such as TabICLv2 or TabPFN-3.
Finally, we note that no benchmark is ever perfect or complete, and neither is ours:
Instead, we embrace the living nature of our benchmark and hope to continuously extend it together with the community -- towards further attention layouts, end-to-end evaluations, and hardware -- as the field of tabular foundation models grows and matures.

\newpage
\bibliographystyle{plainnat}
\bibliography{refs}

\newpage
\appendix
\crefalias{section}{appendix}
\crefalias{subsection}{appendix}
\crefalias{subsubsection}{appendix}
\section{Additional Details}
\label{sec:additional-details}

\subsection{Quick overview of attention backends}
\label{sec:attention-backends}

We briefly summarize the key characteristics of the attention backends evaluated in this work. All implementations compute exact scaled dot-product attention but differ in kernel design, hardware specialization, and support for tensor layouts.

\custompar{SDPA (efficient)} PyTorch's default memory-efficient attention backend, based on xFormers-style kernels~\cite{xFormers2022}.
It reduces memory usage via tiling and recomputation, performs well for small to moderate sequence lengths, and has low kernel launch overhead.
However, it expects contiguous inputs and does not natively support arbitrary stride layouts.

\custompar{SDPA (cuDNN)}
A highly optimized backend dispatching to NVIDIA's cuDNN library~\cite{cudnn,chetlur2014cudnn}.
It achieves strong performance for a wide range of sequence lengths, particularly in the medium regime, by leveraging vendor-tuned kernels.
Similar to the efficient backend, it assumes standard contiguous layouts and may incur overhead from required tensor copies.

\custompar{FlashAttention-2 (FA2)~\cite{dao2023flashattention2}}
An optimized attention kernel designed for Ampere GPUs, focusing on improved parallelism and work partitioning compared to earlier FlashAttention versions.
While memory-efficient and exact, its performance is suboptimal on newer architectures (e.g., Hopper) and it assumes relatively restrictive input layouts.

\custompar{FlashAttention-3 (FA3)~\cite{shah2024flashattention3}}
A further optimized implementation targeting Hopper GPUs, with improved kernel scheduling and support for more flexible stride layouts.
In particular, it can operate directly on transposed (non-contiguous) tensors with compact strides, reducing the need for explicit memory copies.
This makes it well-suited for row attention in tabular settings.

\custompar{FlashAttention-4 (FA4)~\cite{zadouri2026flashattention4}}
A recent iteration co-designed with newer hardware (e.g., Blackwell GPUs), emphasizing kernel pipelining and asymmetric scaling.
FA4 is designed around the observation that tensor core throughput on Blackwell GPUs scales faster than shared memory bandwidth and exponential units, a hardware asymmetry that makes non-matmul operations the bottleneck.
FA4 kernels are optimised to minimise shared memory traffic and redundant softmax rescaling.
It imposes stricter constraints on input dimensions (e.g., minimum head dimension) and is primarily optimized for large models and longer sequences, making it less applicable in typical tabular configurations.

\subsection{A Note on Tabular Attention in different Model Families}
\label{sec:attention-families}
Throughout, we study the interleaved 2D attention of TabPFN-v2-, Mitra-, and ConTextTab-like models in a slightly idealized fashion, evaluating standard full bidirectional attention rather than the exact formulation each architecture uses.
In practice, these architectures differ in how they handle the train--test (context--query) split: TabPFN-v2 and Mitra use a cross-attention mechanism, where the row-attention queries correspond to samples from both train and test while the keys and values come from the training context alone, whereas ConTextTab uses a masking-based approach.
As discussed in \cref{sec:experiments}, our full-attention setup is representative of the zero-query limit and close to the single-query inference case.
Extending the evaluation to the cross-attention formulation is a worthwhile direction, but would substantially enlarge the scope: one would need to sweep both context and query size and distinguish the training regime (a large query size to obtain good gradient estimates) from inference (often a single query row, in which case the setup almost reduces to the full-attention scenario we focus on here).
Similarly, assessing effective speedups with a KV cache depends not only on the tensor shapes but also largely on implementation details and the usage scenario at hand.
Nevertheless, we believe our main findings still translate, as our end-to-end validation of TabPFN-v2.6 in \cref{sec:e2e-validation} indicates: the additional asymmetry between key/value and query sequence lengths will shift the performance profiles depending on the context and query lengths, but the qualitative picture remains.

A second family, following the TabICL approach (TabICLv2~\cite{qu2026tabiclv2}, TabPFN-3~\cite{tabpfn3}), departs more substantially from the setting studied here.
Instead of interleaving row and column attention within each block, these architectures perform row- and column-wise attention in separate, multiple stages.
Many of the table-specific characteristics we identify still transfer, as the core asymmetry between row and column attention persists -- long, wide-batch column sequences versus short-batch, long row sequences -- as do the comparably small head dimensions.
What vanishes is the recurring axis permutation and the associated \texttt{.contiguous()} copies:
for such architectures one would effectively benchmark vanilla 1D attention at the respective batch and sequence lengths but without the transposition calls, closely resembling our column-attention benchmark yet also evaluated at row-attention-relevant shapes.
This is technically straightforward but out of scope for our initial presentation, and a worthwhile extension as an increasing number of models adopt this design.
We note, however, that interleaved 2D attention remains competitive in several settings -- for instance, TabPFN-v2.6 outperforming TabICLv2 on wide tables~\cite{purucker2026beyond} -- so both families remain of practical relevance.

Finally, we note that frameworks such as FlexAttention are, at present, of limited relevance to the challenges studied here.
FlexAttention is a Python-level framework for compiling attention kernels with custom score and mask modifications (e.g.\ ALiBi biases, sliding-window or document masking).
Current tabular architectures, however, do not employ such custom attention formulations, and FlexAttention does not target the difficulties specific to tabular attention: the tensor shapes and sequence lengths that differ substantially from language modeling, and the non-contiguous memory layouts arising from attention in alternating directions.

\subsection{Tensor layout considerations}
\label{sec:layout-ablation}

The standard tensor layout $(B, R, C, H, D)$ places the column dimension before the head dimension.
For column attention, the sequence dimension $C$ is already contiguous, so a simple \texttt{.view()} suffices.
For row attention, the rows and columns must be transposed, producing a non-contiguous tensor.

An alternative layout $(B, C, R, H, D)$ would make row attention contiguous but column attention non-contiguous.
Both layouts incur the same theoretical copy cost for their respective non-contiguous direction: the total number of elements copied is identical.
However, the standard $(B, R, C, H, D)$ layout may provide better cache locality for the non-attention parts of the model (\eg the MLP layers that process features independently per sample), since consecutive features of the same sample are stored contiguously.
Investigating just the \texttt{.contiguous()} performance, we were not able to observe any measurable difference, as shown in \cref{tab:contiguous-ablation}.

\begin{table}[!h]
    \centering
    \caption{Microbenchmark of \texttt{.contiguous()} for the row-first layout (Layout~A) $(B,R,C,H,D)$ vs.\ the column-first layout (Layout~B) $(B,C,R,H,D)$ with $B{=}2$, $H{=}8$, $D{=}64$.  The \texttt{transpose(1,2)} call makes the non-contiguous dimension contiguous.  Times in \unit{\micro\second}; bandwidth in \unit{\giga\byte\per\second}.  All ratios are within \num{0.3}\% of unity, confirming that layout choice does not affect copy cost.}
    \label{tab:contiguous-ablation}
    \small
    \begin{tabular}{rr cc cc c}
        \toprule
        & & \multicolumn{2}{c}{Time (\unit{\micro\second})} & \multicolumn{2}{c}{Bandwidth (\unit{\giga\byte\per\second})} & \\
        \cmidrule(lr){3-4} \cmidrule(lr){5-6}
        $R$ & $C$ & Layout A & Layout B & Layout A & Layout B & Ratio A/B \\
        \midrule
        \num{1024} & \num{20}   & 65.0  & 64.9  & 1290.6 & 1293.1 & 1.002 \\
        \num{2048} & \num{50}   & 310.3 & 310.8 & 1351.6 & 1349.4 & 0.998 \\
        \num{4096} & \num{50}   & 619.5 & 619.7 & 1354.1 & 1353.6 & 1.000 \\
        \num{8192} & \num{100}  & 2479.5 & 2481.8 & 1353.3 & 1352.0 & 0.999 \\
        \num{8192} & \num{16}   & 396.5 & 397.6 & 1354.0 & 1350.1 & 0.997 \\
        \num{128}  & \num{1024} & 398.8 & 399.3 & 1346.1 & 1344.4 & 0.999 \\
        \bottomrule
    \end{tabular}
\end{table}

\subsection{Contiguity, views, and memory copies}
\label{sec:tensor-recap}

PyTorch tensors are defined by a pointer to storage together with metadata (shape and strides).
A tensor is \emph{contiguous} if its elements are laid out in memory such that the stride matches a standard row-major layout.
Many high-performance kernels (including SDPA backends) assume or require such a layout.

Operations such as \texttt{.view()} or \texttt{.reshape()} can return zero-copy views when the requested shape is compatible with the existing stride layout.
In contrast, \texttt{.transpose()} or \texttt{.permute()} typically produce non-contiguous tensors, as they only modify strides without rearranging the underlying storage.
For example, a contiguous tensor of shape $(2,3)$ has strides $(3,1)$; after \texttt{.transpose(0,1)} it has shape $(3,2)$ but strides $(1,3)$, meaning elements are no longer laid out contiguously in memory.

The \texttt{.contiguous()} call enforces a contiguous layout by allocating new memory and copying the tensor data.
This is an $O(N)$ operation in the number of elements and can therefore become a non-negligible overhead, particularly when invoked repeatedly in attention pipelines.
As discussed in \cref{sec:tabular-attention}, this effect is most pronounced for row attention, where transposing from a row-first layout $(B, R, C, H, D)$ to $(B \cdot C, H, R, D)$ produces a non-contiguous tensor that cannot be represented as a simple view,
requiring explicit copies for queries, keys, and values.

In contrast, column attention operates along an already contiguous dimension, allowing reshapes such as $(B, R, C, H, D) \rightarrow (B \cdot R, H, C, D)$ to be implemented as zero-copy views,
with only a final \texttt{.contiguous()} call needed after the attention operation.
Backends that support more flexible stride layouts (\eg FlashAttention-3) can further reduce the need for intermediate copies by operating directly on transposed views.

In summary, whether an operation incurs a memory copy depends on stride compatibility rather than the operation type itself.
In tabular attention, the key distinction is that column attention aligns with the underlying memory layout, while row attention typically does not, making \texttt{.contiguous()} calls -- and their associated cost -- an important practical consideration.

To quantify this cost for the analysis in \cref{sec:analysis}, we measure the achieved bandwidth of \texttt{.contiguous()} on the H100 using the exact tensor shapes from the row attention benchmark ($B_{\text{eff}}{=}64$, $H{=}12$, $D{=}64$, varying $N$).
Each measurement is averaged over 100 repetitions with 10 warmup iterations and L2 cache flushing between repetitions.
\cref{tab:copy-bandwidth} reports the measured bandwidth for three consecutive copies (Q, K, V), showing that it increases with tensor size but saturates well below peak HBM bandwidth due to the strided access pattern.

\begin{table}[!h]
    \centering
    \caption{Measured \texttt{.contiguous()} bandwidth for three copies (Q, K, V) on the H100 at the row attention shapes used in the benchmark. Configuration: $B_{\text{eff}}{=}64, H{=}12, N, D{=}64$ in \texttt{bfloat16}.}
    \label{tab:copy-bandwidth}
    \small
    \begin{tabular}{r r r r}
        \toprule
        $N$ & Total size (\unit{\mega\byte}) & Time (\unit{\micro\second}) & Bandwidth (\unit{\giga\byte\per\second}) \\
        \midrule
        \num{32}    & 9     & 24.1   & 783  \\
        \num{64}    & 18    & 33.2   & 1138 \\
        \num{128}   & 36    & 56.8   & 1330 \\
        \num{256}   & 72    & 115.4  & 1308 \\
        \num{512}   & 144   & 220.5  & 1370 \\
        \num{1024}  & 288   & 422.7  & 1429 \\
        \num{2048}  & 576   & 830.6  & 1454 \\
        \num{4096}  & 1152  & 1646.0 & 1468 \\
        \num{8192}  & 2304  & 3307.4 & 1461 \\
        \num{16384} & 4608  & 6665.1 & 1450 \\
        \bottomrule
    \end{tabular}
\end{table}

\newpage
\section{Additional Results}
\label{sec:additional-results}

In this appendix, we provide the relative speedup plot and supplementary results for the main head count and dimension present in current tabular foundation models, showing both backward and forward path throughput and peak memory across the investigated GPUs.
Due to the large number of configurations evaluated, we only present a selective subset here.
Note that evaluation plots for all configurations across all benchmarked GPU families can be generated from the public benchmark repository, which contains all raw measurement results.

\subsection{Batch Size Sweep}
\begin{figure}[t]
    \centering
    \includegraphics[width=\textwidth,trim={0 6mm 0 14mm},clip]{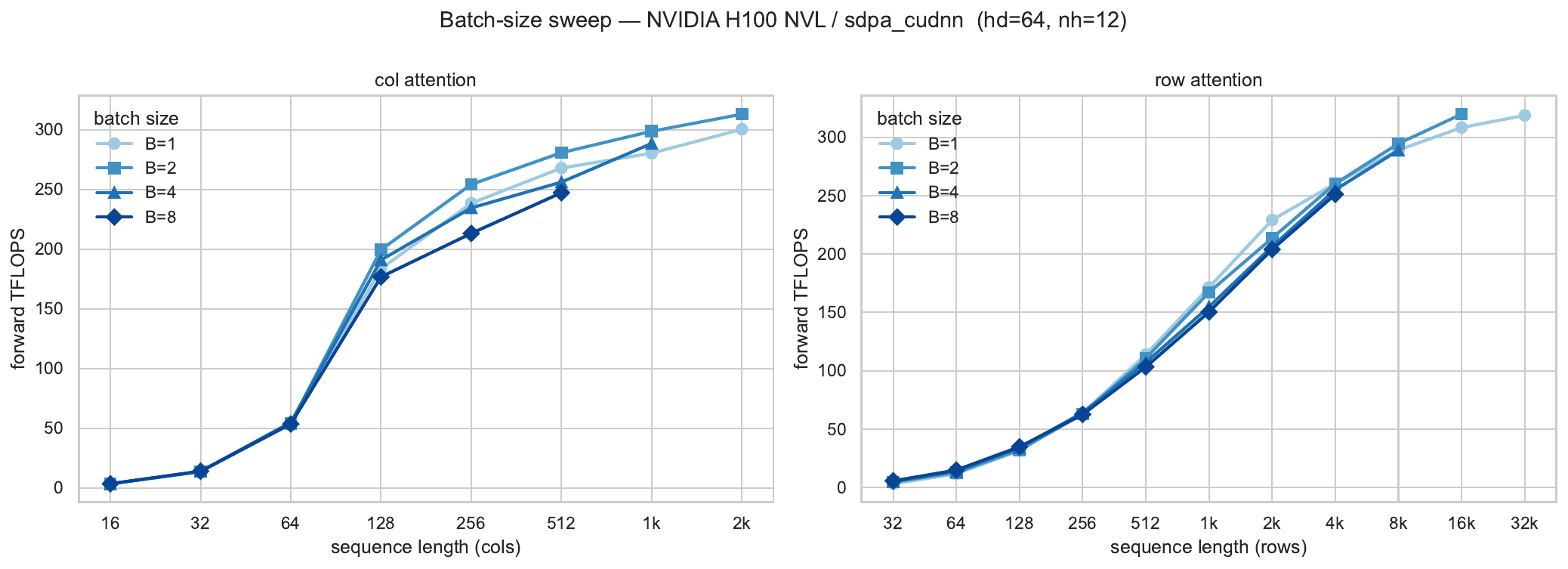}
    \caption{H100 column attention (left) and row attention (right) throughput for different batch sizes for $H{=}\num{12}$ heads with head dimension $D{=}\num{64}$.}
    \label{fig:h100-batch-ablation}
\end{figure}

To validate that our $B{=}1$ setup is not too restrictive, we evaluate throughput for batch sizes 1, 2, 4, and 8 for the H100 in the case of 12 heads and 64 head dimension.
The results are shown in \cref{fig:h100-batch-ablation}.
We can observe what we expected: the varying batch size basically multiplies the effective batch size given by the column or row number for row and column attention, respectively.
As such we see little difference between different batch sizes in the case of row attention, where the overall effective batch size is comparably small.
For column attention, we see qualitatively equivalent scaling for the different configurations, with slightly reduced throughput as the batch size grows in the long-sequence regime, while we do not observe any differences for smaller column counts.

\subsection{Peak Memory}
\begin{figure}[t]
    \centering
    \includegraphics[width=\textwidth,trim={0 6mm 0 14mm},clip]{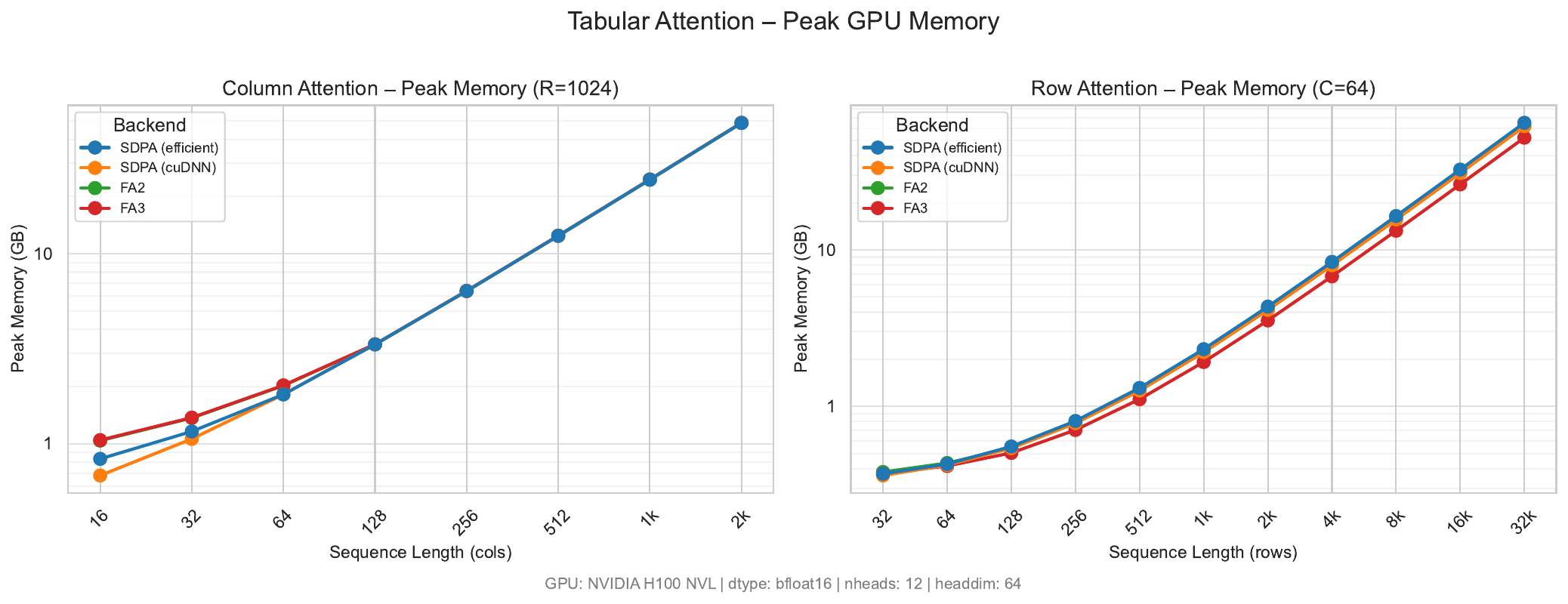}
    \caption{H100 column attention (left) and row attention (right) peak memory for $H{=}\num{12}$ heads with head dimension $D{=}\num{64}$.}
    \label{fig:h100-peak-memory}
\end{figure}

To illustrate the difference in memory consumption, we show the peak memory in the case of column and row attention for the H100 in the case of 12 heads with 64 dimensions, but note that results for other configurations are conceptually identical, however with the ability to scale to longer row sequences as we run out of CUDA memory for configurations with larger hidden dimension naturally.
The result is depicted in \cref{fig:h100-peak-memory}.
As expected we see basically identical scaling behaviour of the different backends (which all have $\mathcal{O}(N)$ memory complexity) with only minor absolute differences and FA3 showing a slightly less overall footprint as compared to the remaining backends.
As a result, however, backends basically run out of memory at identical orders of magnitude.
Hence, the backend choice can be made regardless of memory constraints as no backend has a clear advantage over the other in terms of peak memory.

\subsection{Throughput}

\custompar{Hidden dimension \num{128} with \num{8} heads at \num{16} (TabICL setup)}

\begin{figure}[t]
    \centering
    \includegraphics[width=\textwidth,trim={0 6mm 0 14mm},clip]{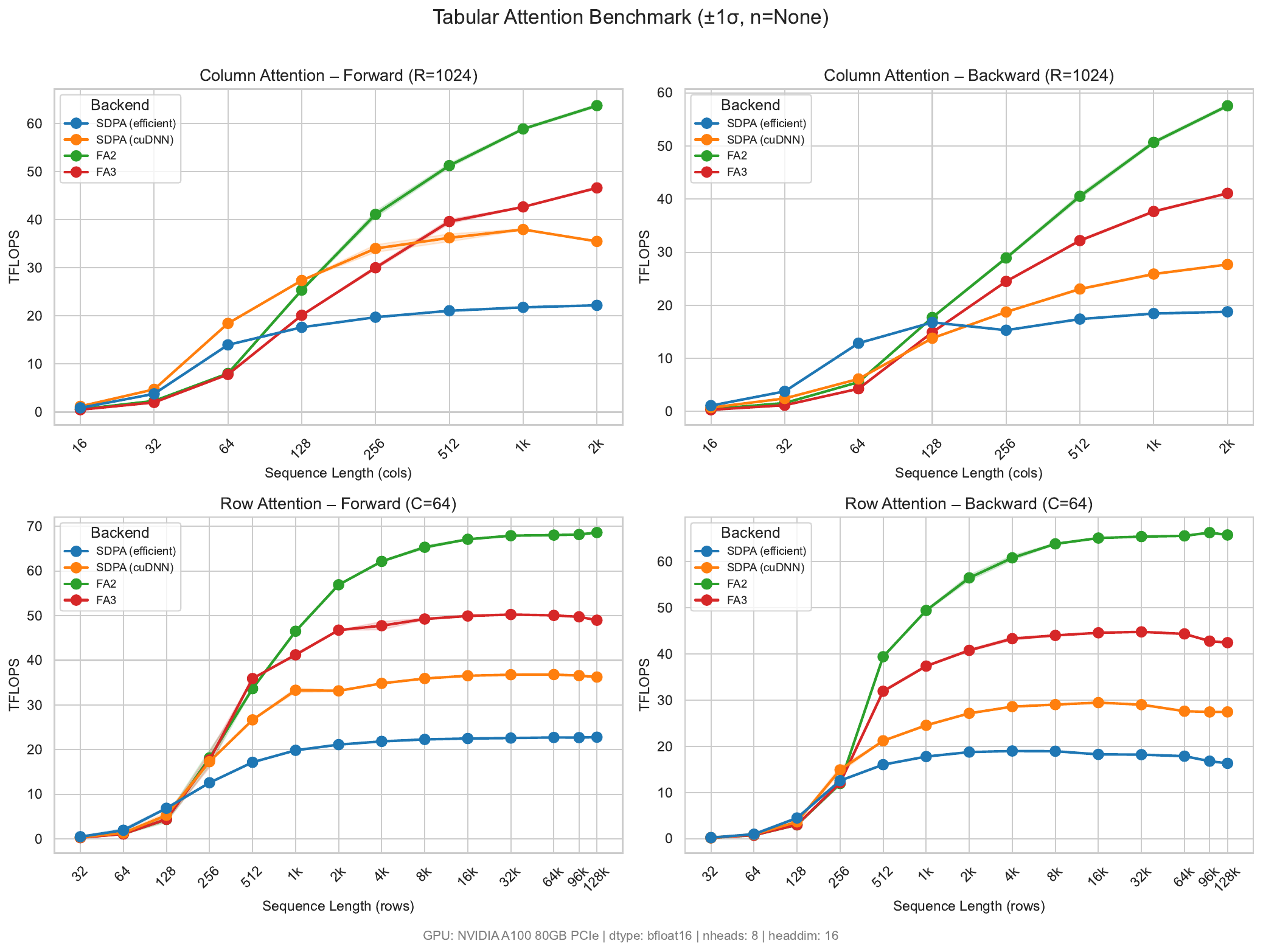}
    \includegraphics[width=\textwidth,trim={0 6mm 0 14mm},clip]{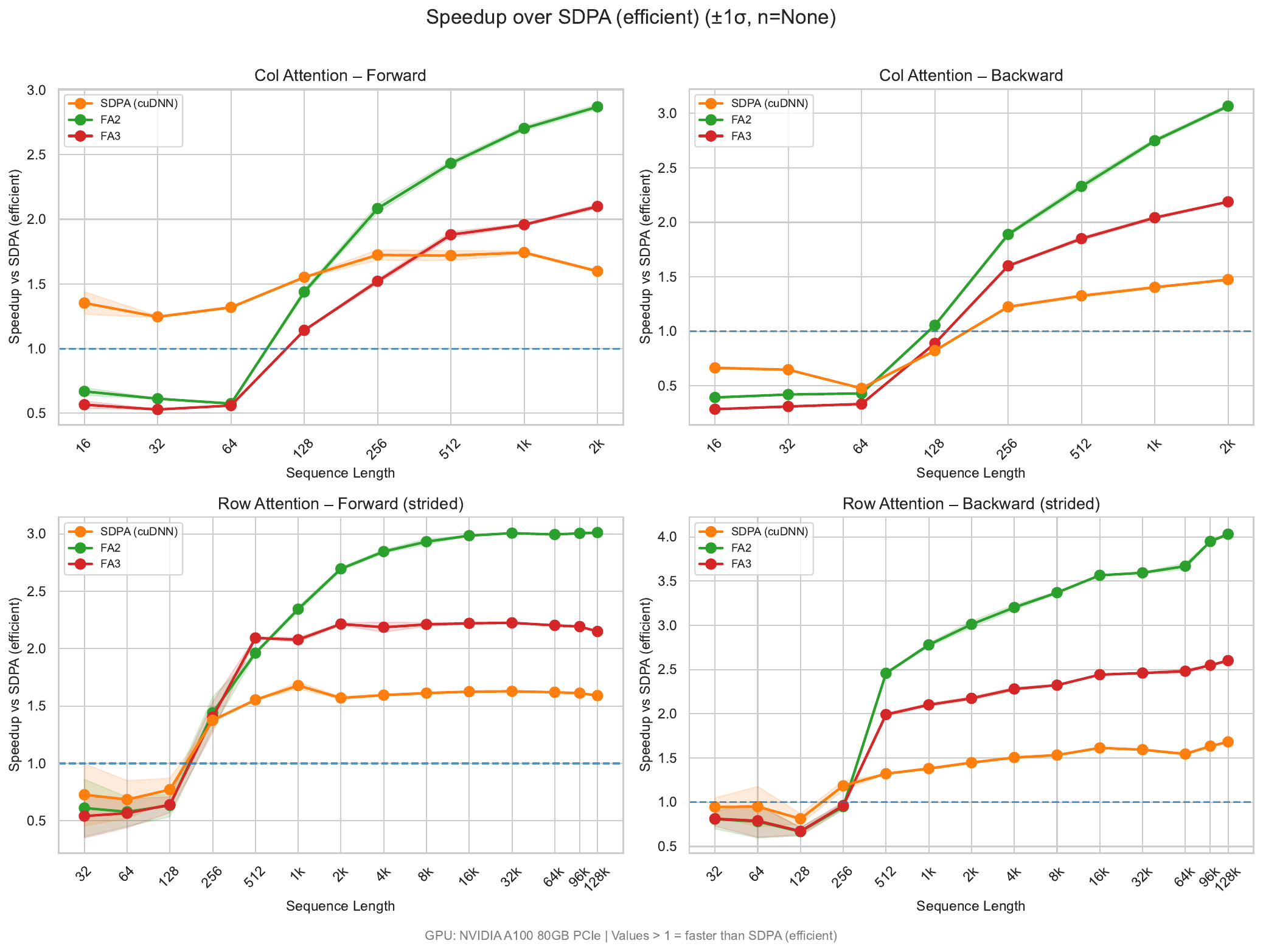}
    \caption{A100 throughput (top) and speedup (bottom) of each backend relative to the SDPA (efficient) baseline in the case of a hidden dimension of $d{=}\num{128}$ consisting of $H{=}\num{8}$ heads with head dimension $D{=}\num{16}$. Error bars show $\pm 1\sigma$ over \num{50} repetitions.}
    \label{fig:a100-128-h8-d16}
\end{figure}

\begin{figure}[t]
    \centering
    \includegraphics[width=\textwidth,trim={0 6mm 0 14mm},clip]{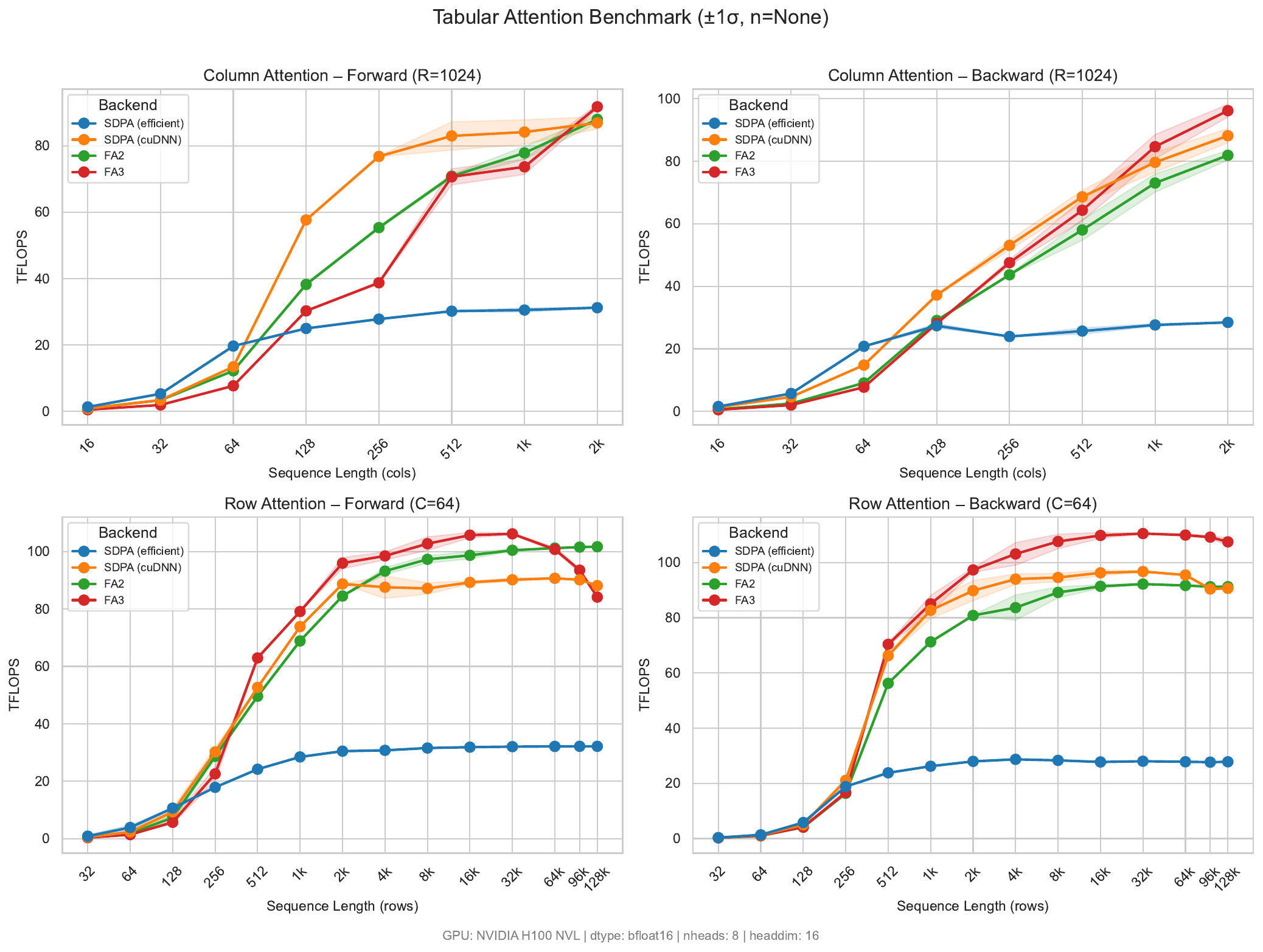}
    \includegraphics[width=\textwidth,trim={0 6mm 0 14mm},clip]{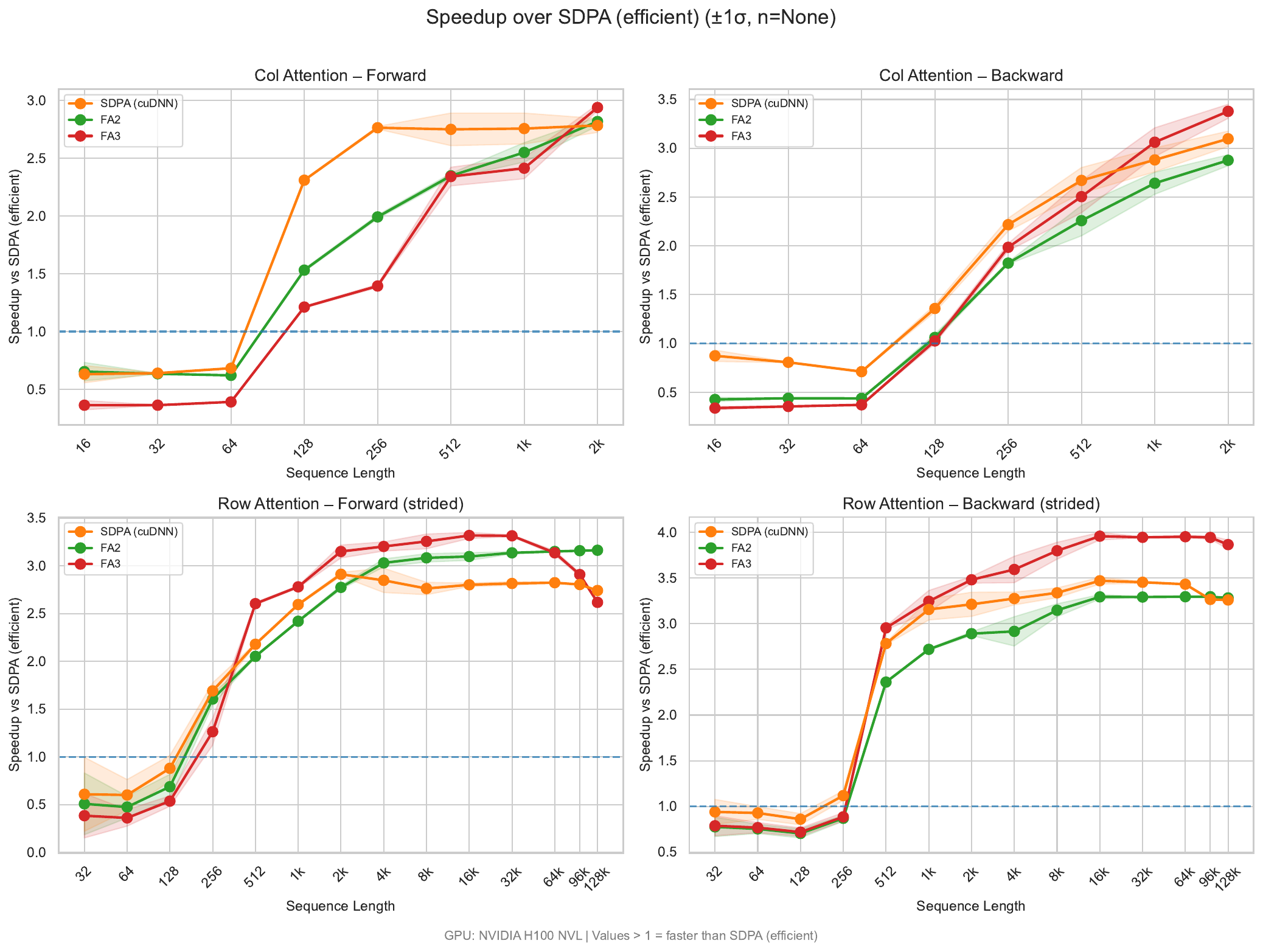}
    \caption{H100 throughput (top) and speedup (bottom) of each backend relative to the SDPA (efficient) baseline in the case of a hidden dimension of $d{=}\num{128}$ consisting of $H{=}\num{8}$ heads with head dimension $D{=}\num{16}$. Error bars show $\pm 1\sigma$ over \num{50} repetitions.}
    \label{fig:h100-128-h8-d16}
\end{figure}

For the A100, the results are shown in \cref{fig:a100-128-h8-d16}.\\
For the H100, the results are shown in \cref{fig:h100-128-h8-d16}. \\
Due to the lack of FlashAttention-4 support on the B200 for a head dimension of 16, we did not evaluate this setting on B200 for the remaining backends.

\custompar{Hidden dimension \num{192} with \num{6} heads at \num{32} (TabPFN setup)}

\begin{figure}[t]
    \centering
    \includegraphics[width=\textwidth,trim={0 6mm 0 14mm},clip]{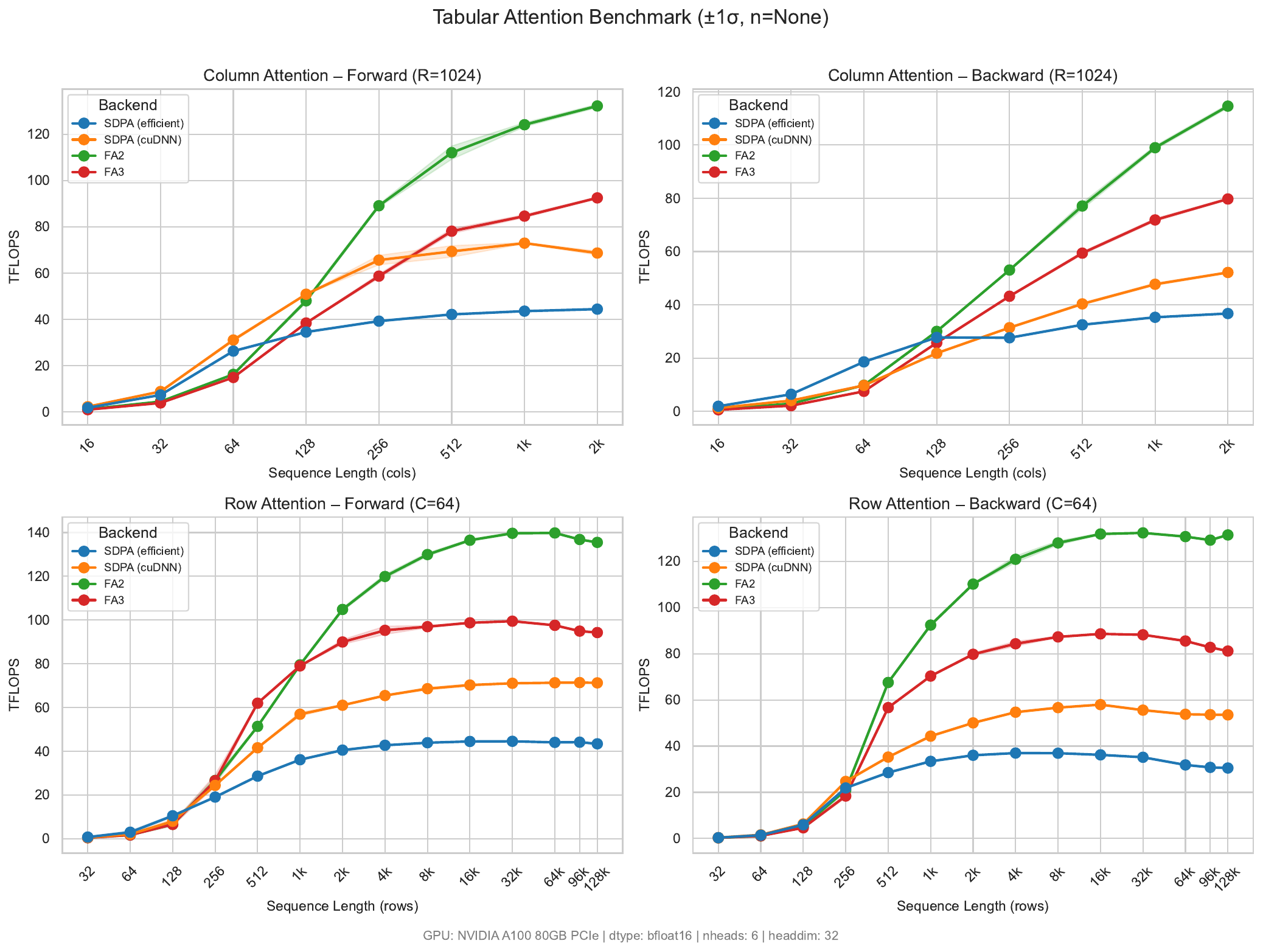}
    \includegraphics[width=\textwidth,trim={0 6mm 0 14mm},clip]{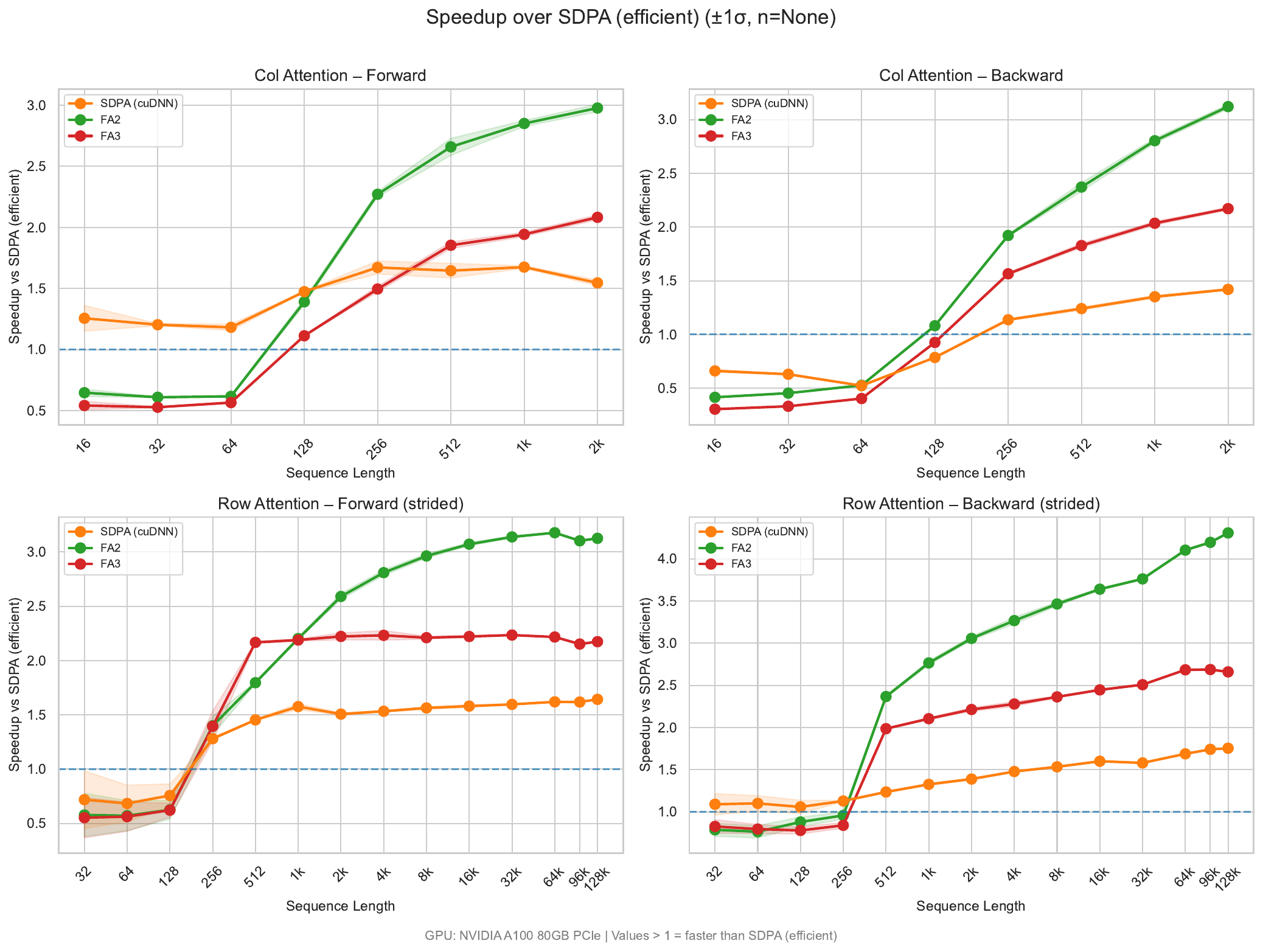}
    \caption{A100 throughput (top) and speedup (bottom) of each backend relative to the SDPA (efficient) baseline in the case of a hidden dimension of $d{=}\num{192}$ consisting of $H{=}\num{6}$ heads with head dimension $D{=}\num{32}$. Error bars show $\pm 1\sigma$ over \num{50} repetitions.}
    \label{fig:a100-192-h6-d32}
\end{figure}

\begin{figure}[t]
    \centering
    \includegraphics[width=\textwidth,trim={0 6mm 0 14mm},clip]{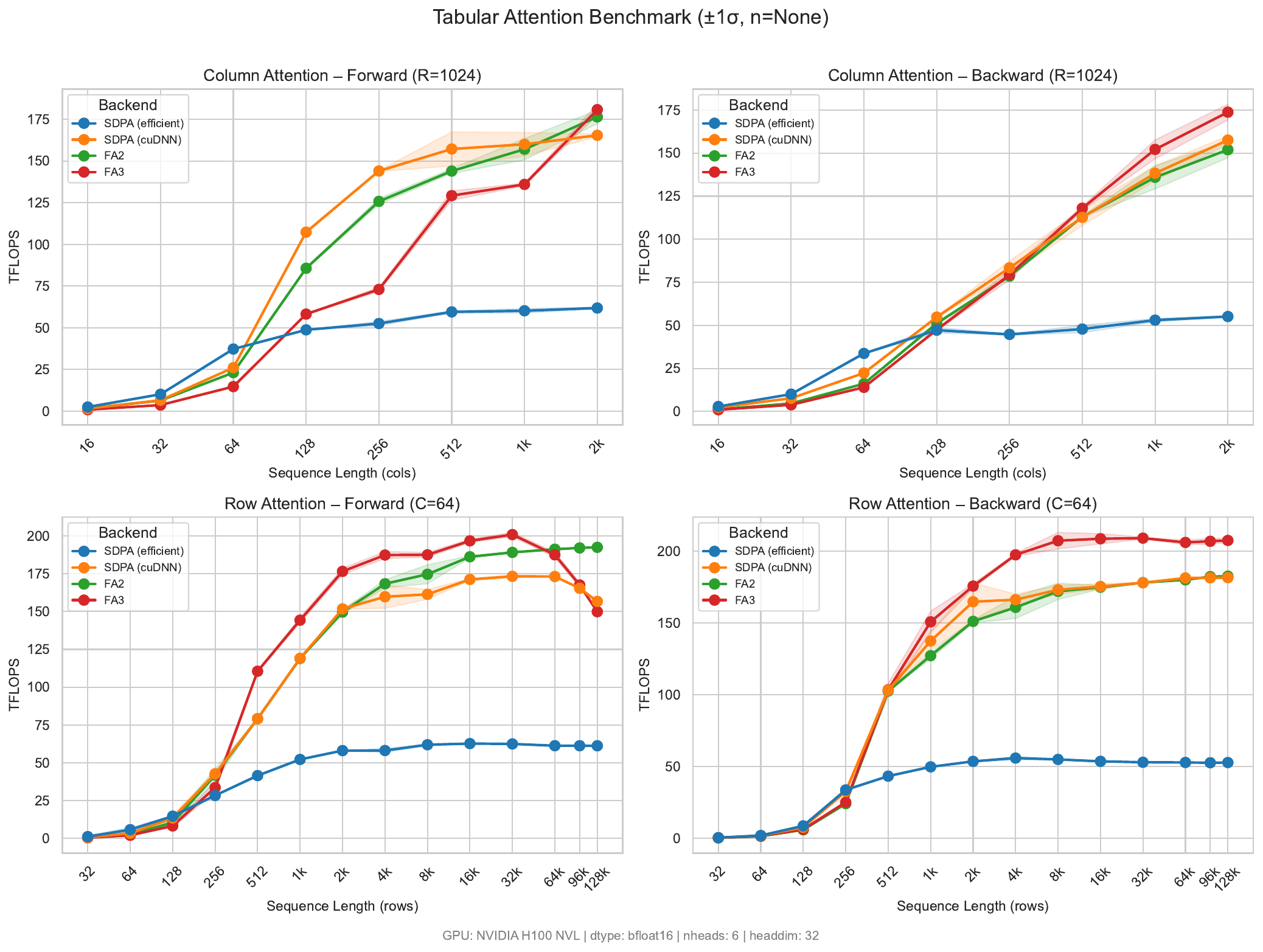}
    \includegraphics[width=\textwidth,trim={0 6mm 0 14mm},clip]{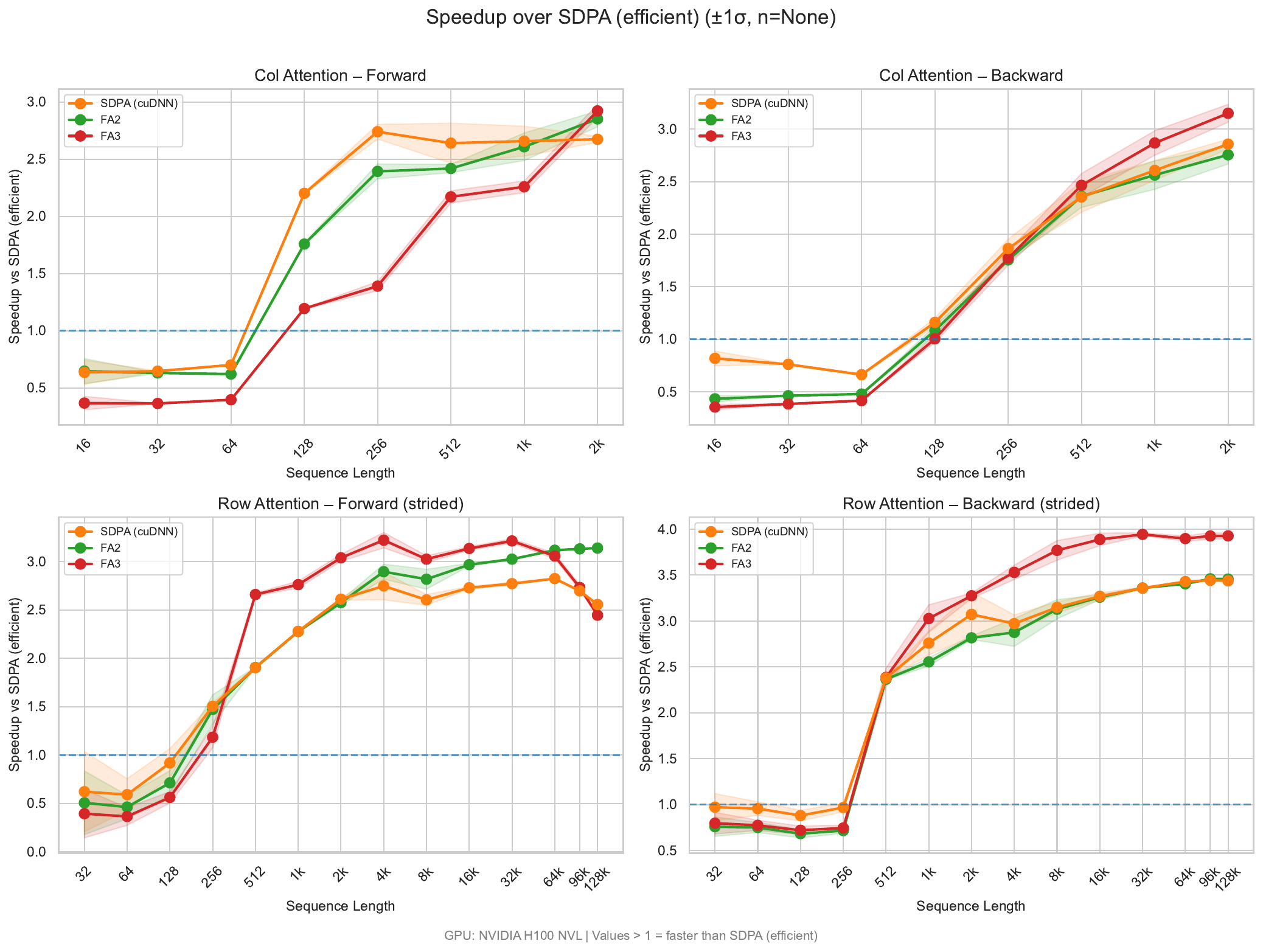}
    \caption{H100 throughput (top) and speedup (bottom) of each backend relative to the SDPA (efficient) baseline in the case of a hidden dimension of $d{=}\num{192}$ consisting of $H{=}\num{6}$ heads with head dimension $D{=}\num{32}$. Error bars show $\pm 1\sigma$ over \num{50} repetitions.}
    \label{fig:h100-192-h6-d32}
\end{figure}

\begin{figure}[t]
    \centering
    \includegraphics[width=\textwidth,trim={0 6mm 0 14mm},clip]{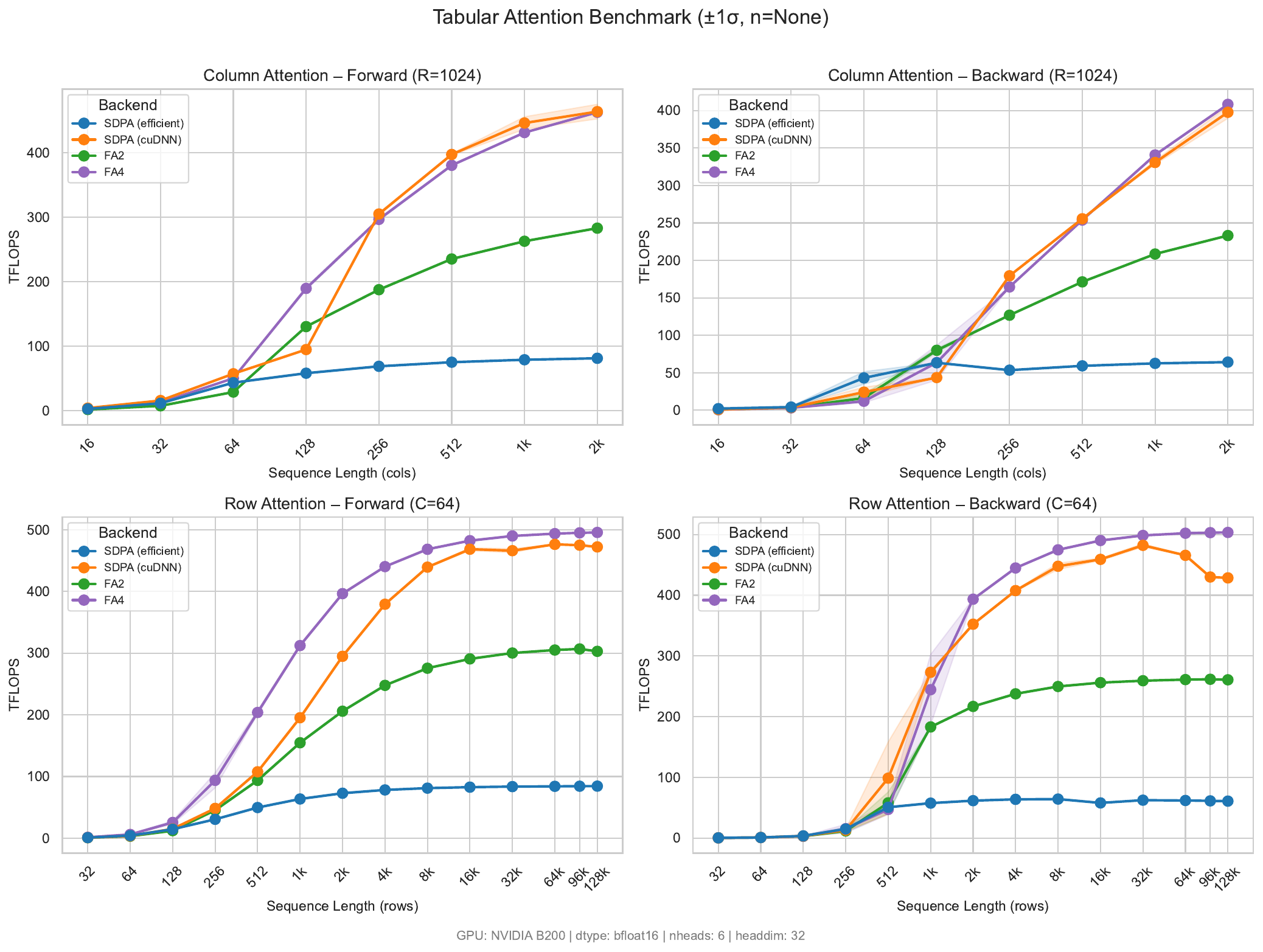}
    \includegraphics[width=\textwidth,trim={0 6mm 0 14mm},clip]{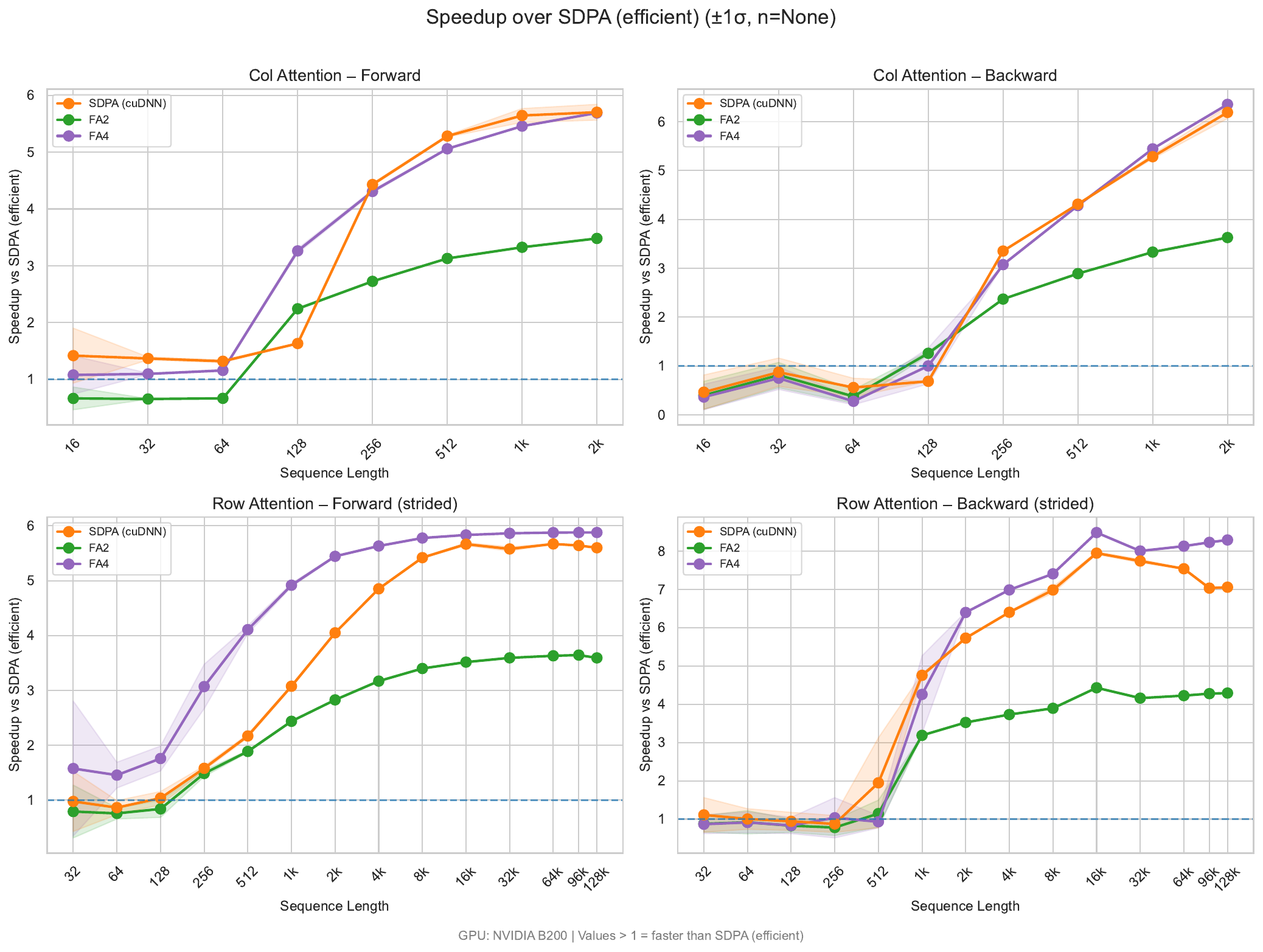}
    \caption{B200 throughput (top) and speedup (bottom) of each backend relative to the SDPA (efficient) baseline in the case of a hidden dimension of $d{=}\num{192}$ consisting of $H{=}\num{6}$ heads with head dimension $D{=}\num{32}$. Error bars show $\pm 1\sigma$ over \num{50} repetitions.}
    \label{fig:b200-192-h6-d32}
\end{figure}

For the A100, the results are shown in \cref{fig:a100-192-h6-d32}.\\
For the H100, the results are shown in \cref{fig:h100-192-h6-d32}. \\
For the B200, the results are shown in \cref{fig:b200-192-h6-d32}.\\

\custompar{Hidden dimension \num{768} with \num{12} heads at \num{64} (ConTextTab setup)}

\begin{figure}[t]
    \centering
    \includegraphics[width=\textwidth,trim={0 6mm 0 14mm},clip]{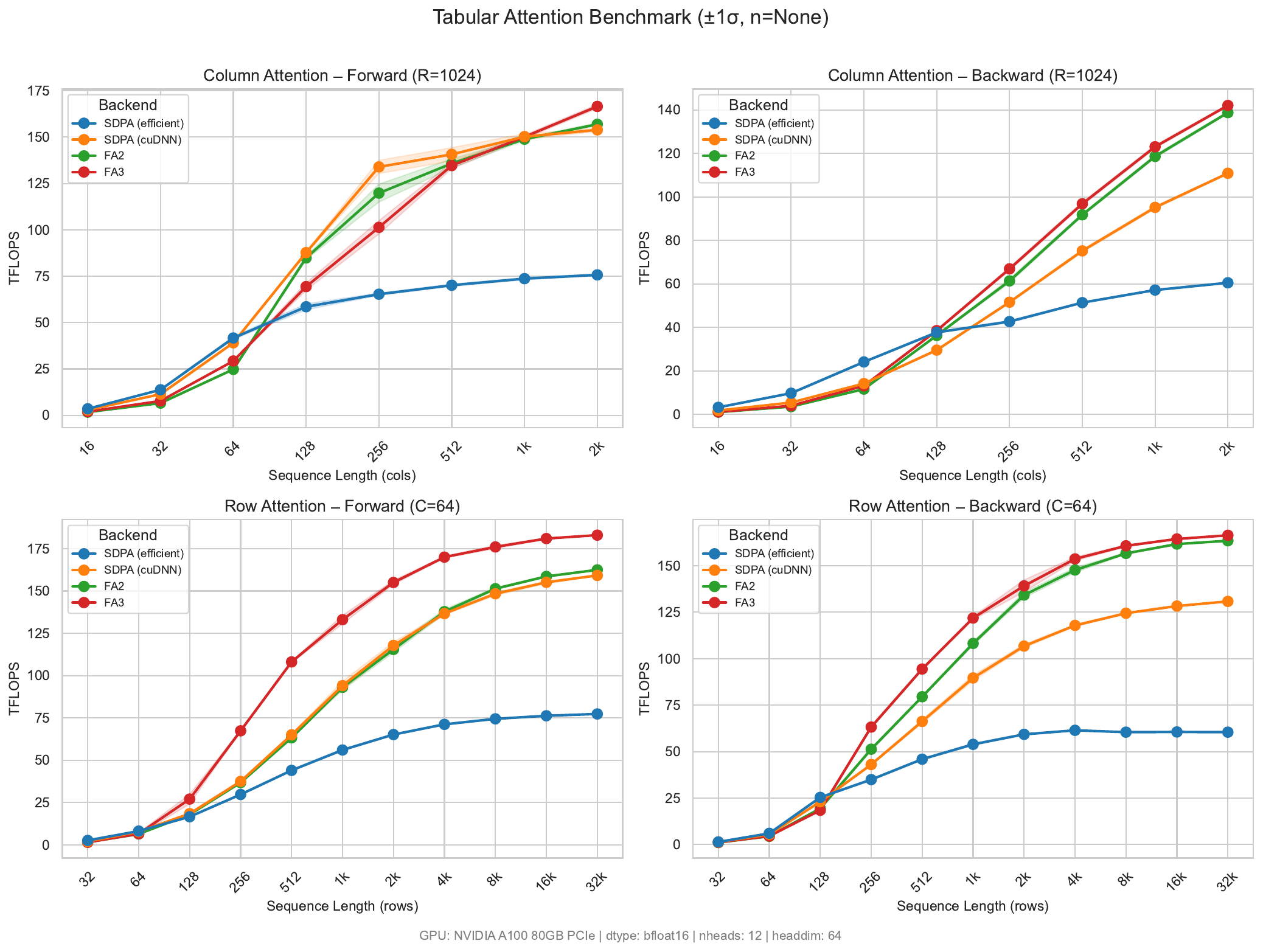}
    \includegraphics[width=\textwidth,trim={0 6mm 0 14mm},clip]{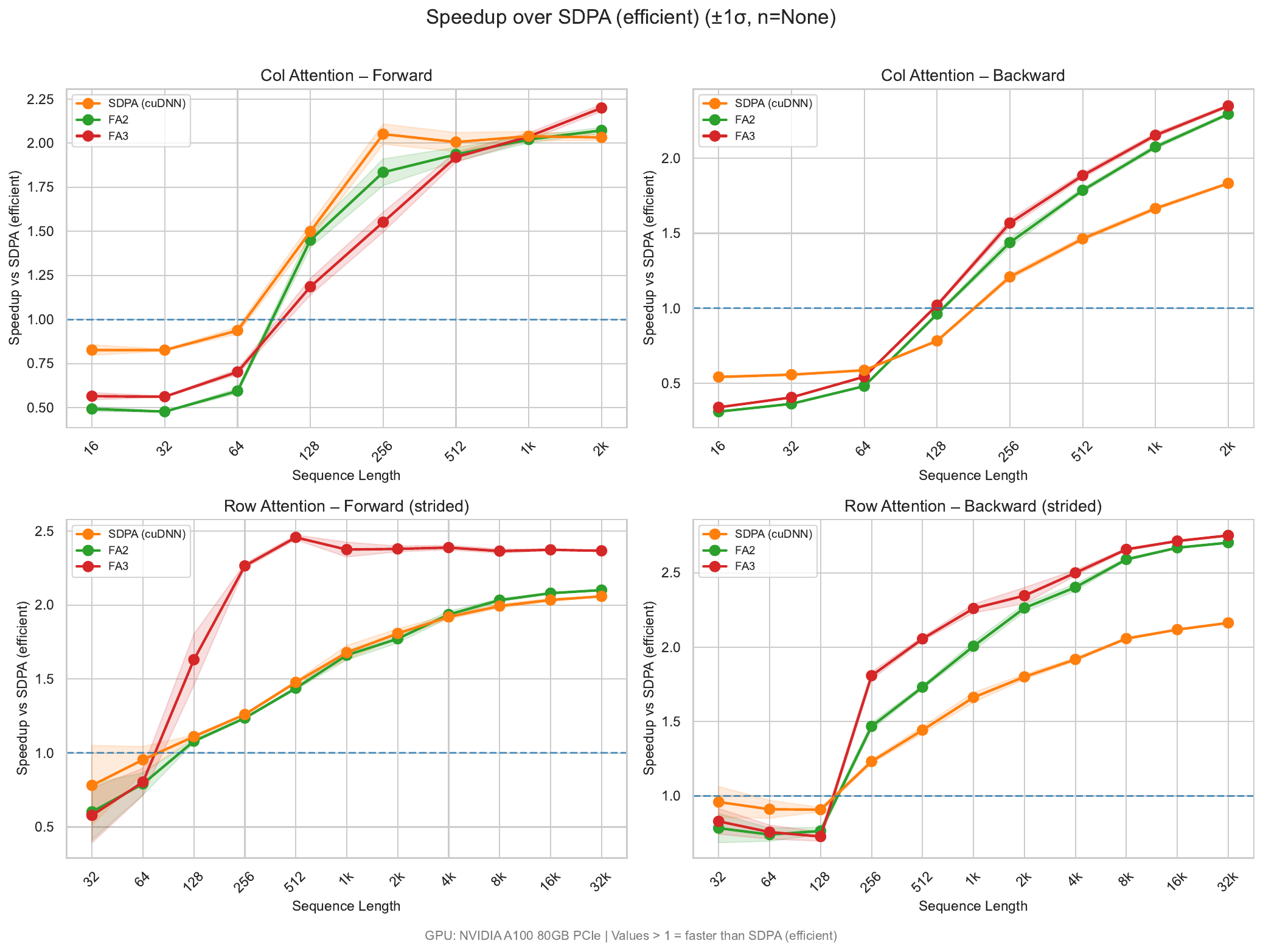}
    \caption{A100 throughput (top) and speedup (bottom) of each backend relative to the SDPA (efficient) baseline in the case of a hidden dimension of $d{=}\num{768}$ consisting of $H{=}\num{12}$ heads with head dimension $D{=}\num{64}$. Error bars show $\pm 1\sigma$ over \num{50} repetitions.}
    \label{fig:a100-768-h12-d64}
\end{figure}

\begin{figure}[t]
    \centering
    \includegraphics[width=\textwidth,trim={0 6mm 0 14mm},clip]{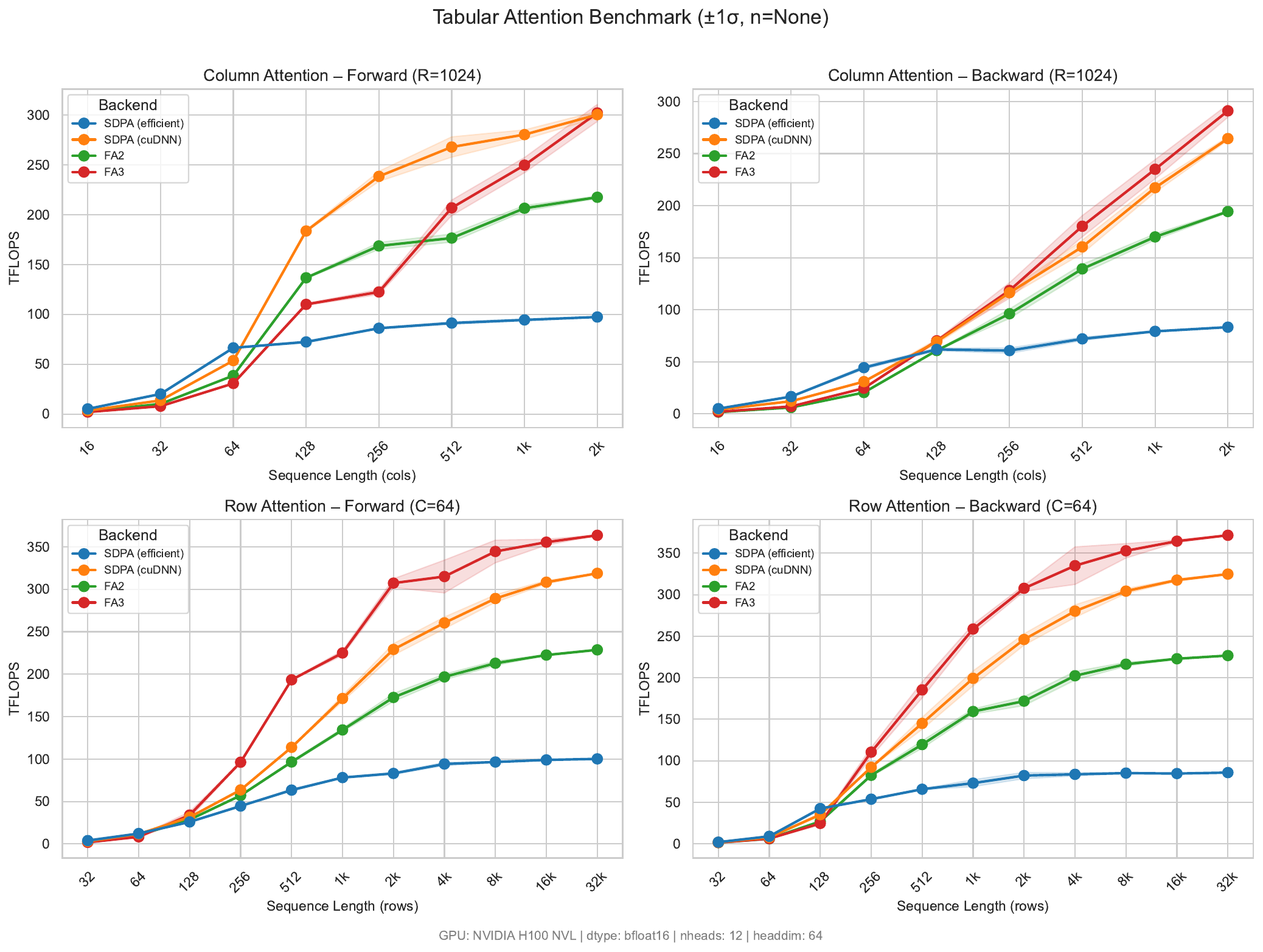}
    \includegraphics[width=\textwidth,trim={0 6mm 0 14mm},clip]{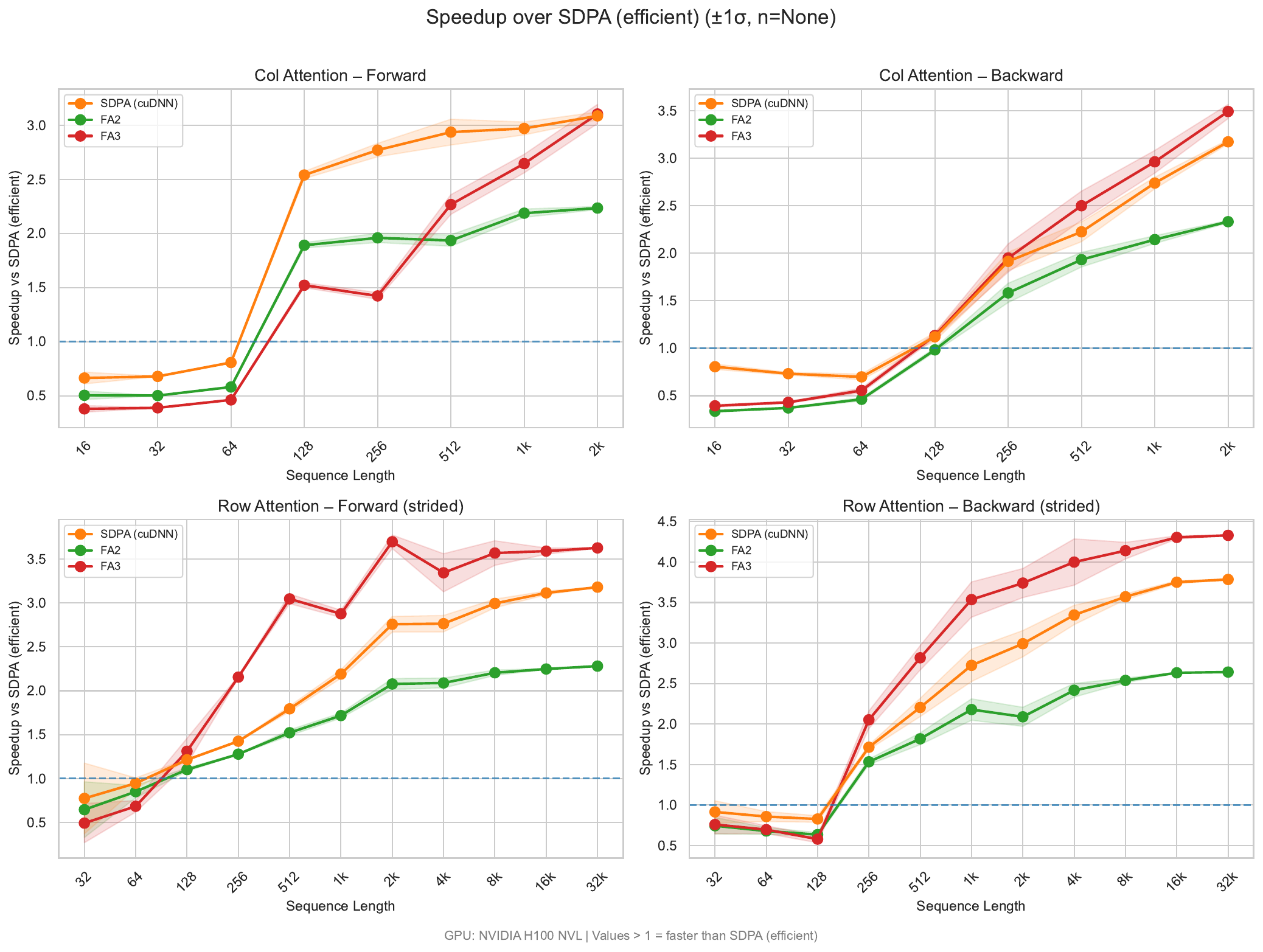}
    \caption{H100 throughput (top) and speedup (bottom) of each backend relative to the SDPA (efficient) baseline in the case of a hidden dimension of $d{=}\num{768}$ consisting of $H{=}\num{12}$ heads with head dimension $D{=}\num{64}$. Error bars show $\pm 1\sigma$ over \num{50} repetitions.}
    \label{fig:h100-768-h12-d64}
\end{figure}

\begin{figure}[t]
    \centering
    \includegraphics[width=\textwidth,trim={0 6mm 0 14mm},clip]{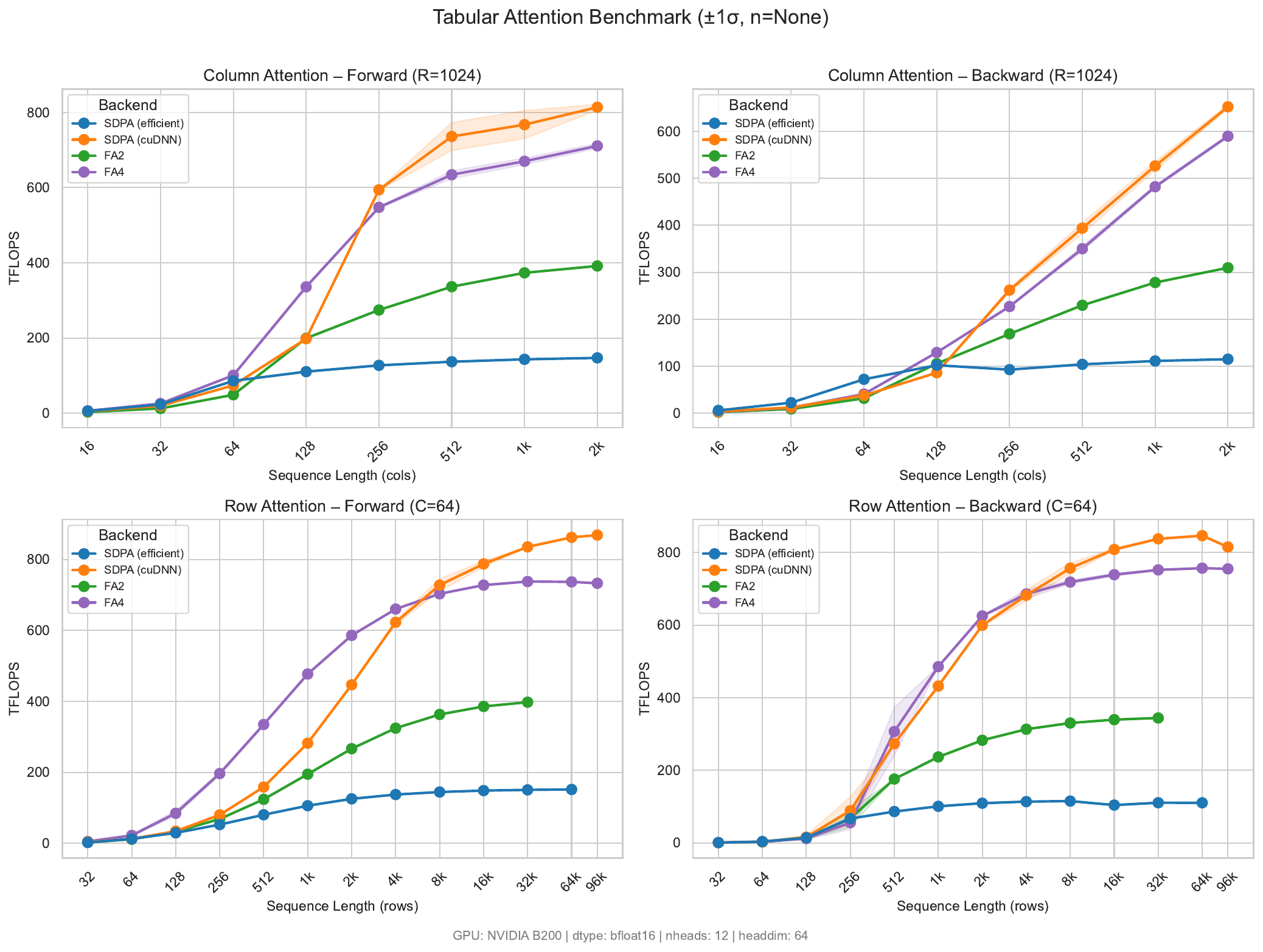}
    \includegraphics[width=\textwidth,trim={0 6mm 0 14mm},clip]{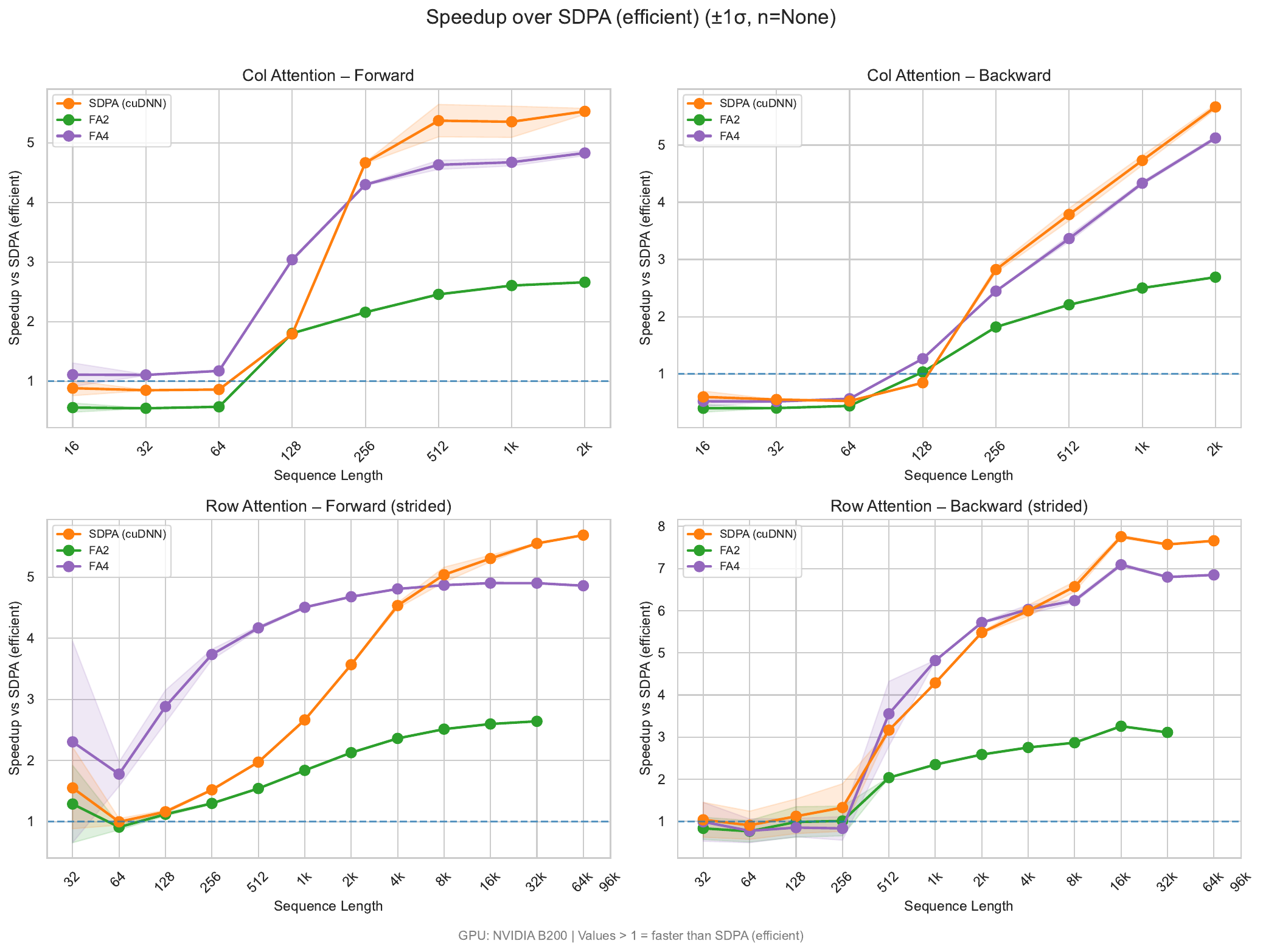}
    \caption{B200 throughput (top) and speedup (bottom) of each backend relative to the SDPA (efficient) baseline in the case of a hidden dimension of $d{=}\num{768}$ consisting of $H{=}\num{12}$ heads with head dimension $D{=}\num{64}$. Error bars show $\pm 1\sigma$ over \num{50} repetitions.}
    \label{fig:b200-768-h12-d64}
\end{figure}

For the A100, the results are shown in \cref{fig:a100-768-h12-d64}.\\
For the H100, the results are shown in \cref{fig:h100-768-h12-d64}. \\
For the B200, the results are shown in \cref{fig:b200-768-h12-d64}. \\




\subsection{End-to-end throughput measurements on TabArena}
\begin{figure}[t]
    \centering
    \begin{subfigure}[b]{0.38\textwidth}
        \centering
        \includegraphics[width=\textwidth,trim={0 2mm 0 11mm},clip]{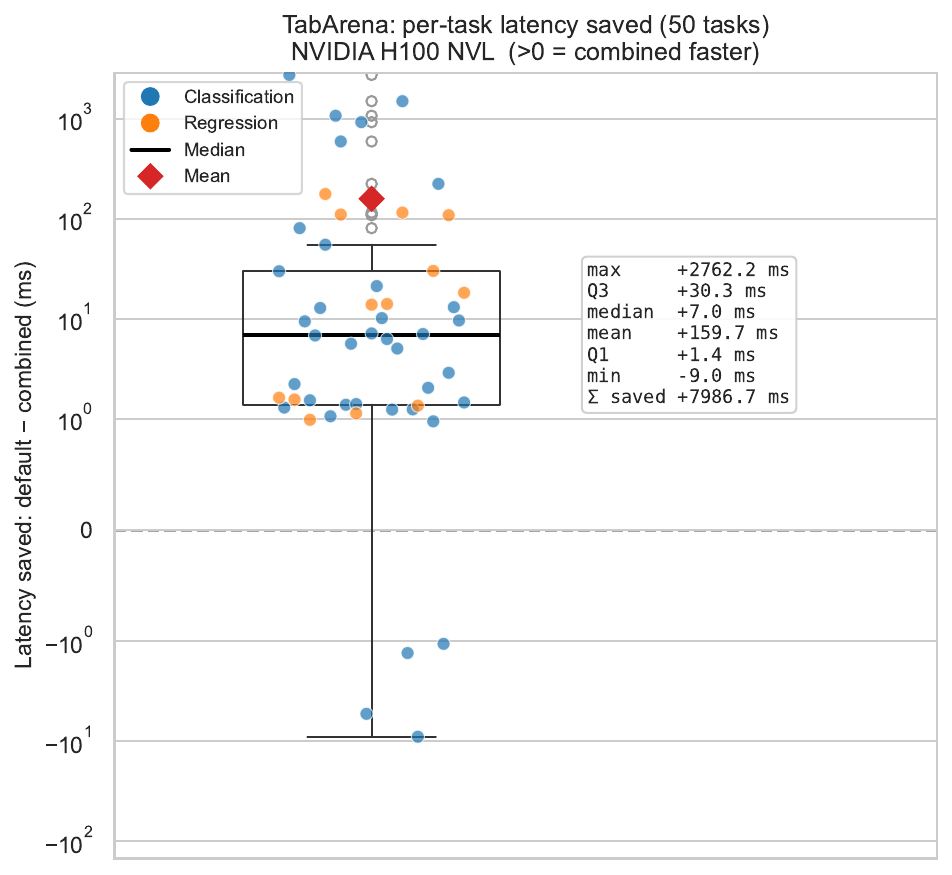}
        \caption{Latency difference distribution.}
        \label{fig:e2e-tabarena-latency-box}
    \end{subfigure}
    \hfill
    \begin{subfigure}[b]{0.58\textwidth}
        \centering
        \includegraphics[width=\textwidth,trim={0 2mm 0 12mm},clip]{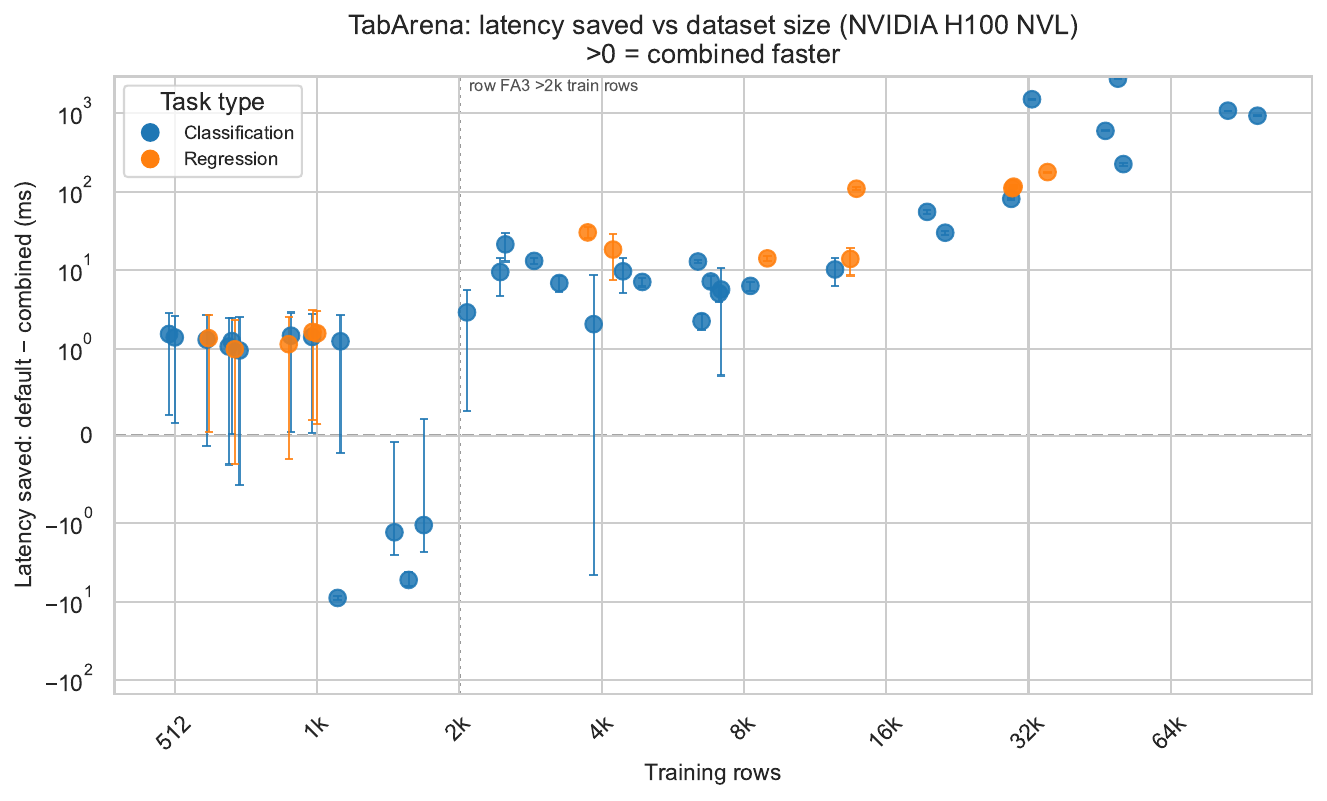}
        \caption{Latency difference versus dataset size.}
        \label{fig:e2e-tabarena-latency-size}
    \end{subfigure}\\[2mm]
    \begin{subfigure}[b]{0.38\textwidth}
        \centering
        \includegraphics[width=\textwidth,trim={0 1mm 0 11mm},clip]{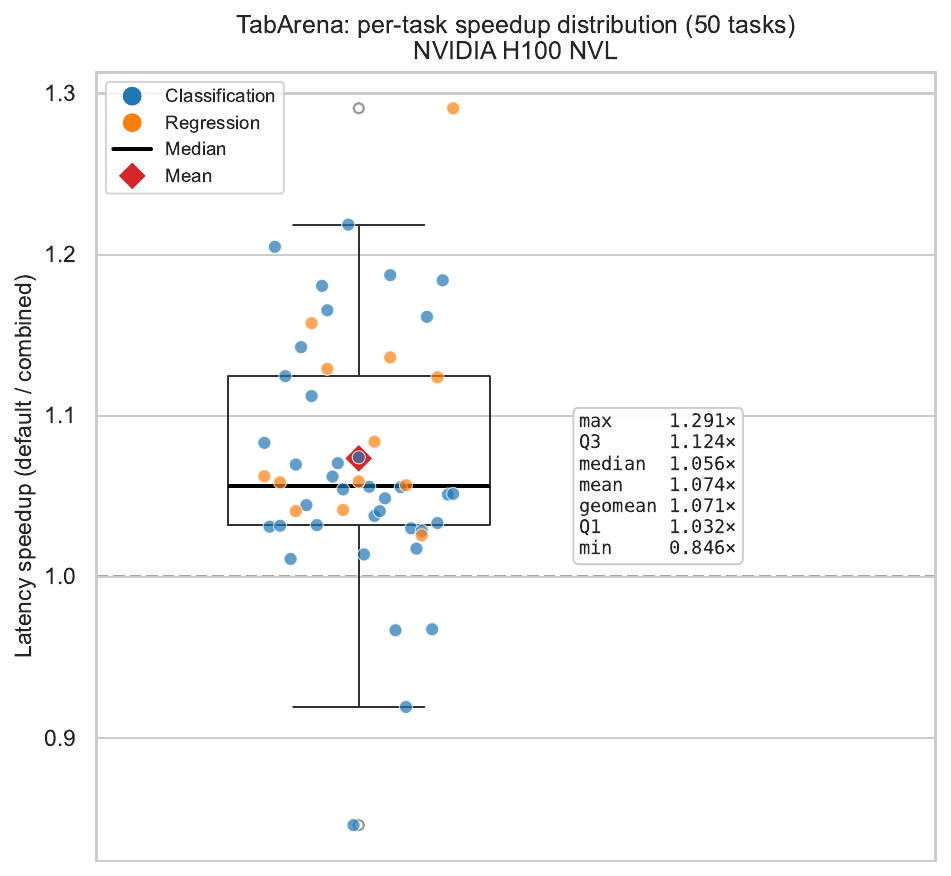}
        \caption{Speedup distribution.}
        \label{fig:e2e-tabarena-speedup-box}
    \end{subfigure}
    \hfill
    \begin{subfigure}[b]{0.58\textwidth}
        \centering
        \includegraphics[width=\textwidth,trim={0 1mm 0 12mm},clip]{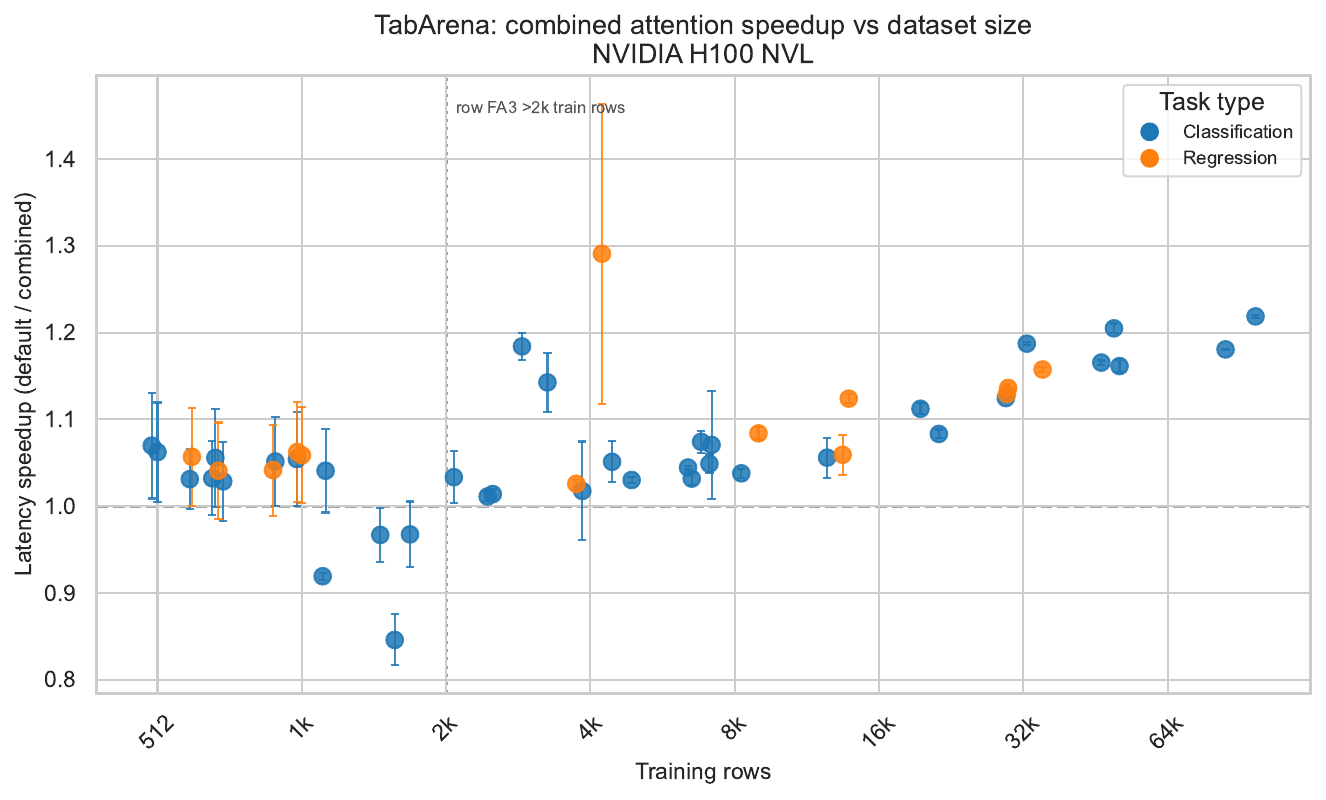}
        \caption{Speedup versus dataset size.}
        \label{fig:e2e-tabarena-speedup-size}
    \end{subfigure}
    \caption{End-to-end latency improvements (top) and speedup (bottom) of the optimized mixed-backend dispatch relative to TabPFN-v2.6's default attention across all TabArena-lite tasks on the H100 GPU.
    The left column shows the distribution over tasks; the right column shows the dependence on dataset size.}
    \label{fig:e2e-tabarena}
\end{figure}

As denoted in \cref{sec:e2e-validation}, we present additional results on the end-to-end evaluation of TabPFN-v2.6 on the TabArena benchmarks.
The results are shown in \cref{fig:e2e-tabarena}, comparing the mixed-backend dispatch against TabPFN-v2.6's default attention across all TabArena-lite tasks on the H100 GPU.
The combined dispatch is faster on the large majority of tasks (\qty{92}{\percent}), with a mean speedup of \qty{7.4}{\percent} and a maximum of \qty{29}{\percent}, translating to absolute latency reductions of up to roughly \qty{2.8}{\second} on the largest tasks.
On a handful of small tasks the mixed dispatch is marginally slower, as the per-call kernel-selection overhead is no longer amortized when the attention itself is cheap.

As the right column of \cref{fig:e2e-tabarena} shows, the speedup scales clearly with dataset size:
the gains are modest on small tables (a mean of \qty{4.5}{\percent} for tasks below \num{10000} training rows) but grow substantially on the larger ones (a mean of \qty{14}{\percent} above \num{10000} rows), consistent with the isolated row-attention results where the FlashAttention backends only overtake the baselines at longer sequence lengths.
Importantly, the predictive scores are essentially unchanged between the two dispatches (a maximum absolute difference of \num{0.004} across all tasks, evaluating accuracy for classification and $R^2$ for regression tasks), confirming that the speedups come at no cost to predictive performance.

\subsection{End-to-end validation of SageAttention}
We have previously seen promising speedup results of SageAttention in the case of long sequence lengths.
What remains unclear is what impact this approximate attention kernel has on the predictive performance.
This, of course, is model dependent.
To illustrate its effect, we evaluate the impact of SageAttention on the predictive performance of TabPFN-v2.6~\cite{tabpfnv2} as well as ConTextTab~\cite{contexttab} across the classification tasks of the CARTE and TabArena-lite benchmarks~\cite{CARTE,tabarena}, covering a total of \num{48} tasks ranging from \num{500} to \num{100000} rows and between \num{4} and \num{1700} columns.
Throughout, we evaluate accuracy as the target metric and compare each SageAttention-based model against its respective base implementation.
To assess statistical significance, we perform a Friedman test with a post-hoc Nemenyi test using the \texttt{autorank}~\cite{autorank} library, from which we obtain the critical-difference diagrams reported below.

\begin{figure}[t]
    \centering
    \begin{subfigure}[b]{0.9\textwidth}
        \centering
        \includegraphics[width=\textwidth,trim={12mm 5mm 12mm 2mm},clip]{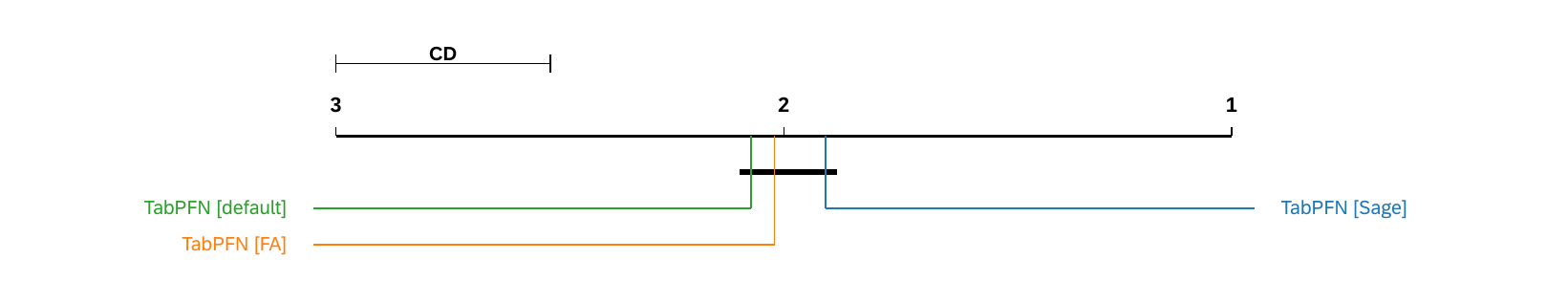}
        \caption{TabPFN.}
        \label{fig:e2e-tabpfn-sage}
    \end{subfigure}\\[1mm]
    \begin{subfigure}[b]{0.9\textwidth}
        \centering
        \includegraphics[width=\textwidth,trim={12mm 5mm 12mm 2mm},clip]{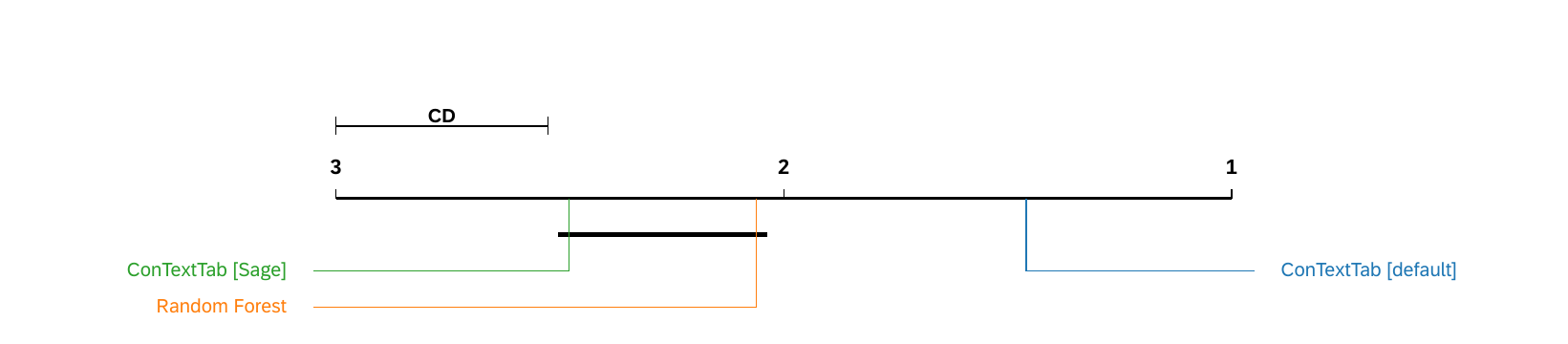}
        \caption{ConTextTab.}
        \label{fig:e2e-contexttab-sage}
    \end{subfigure}
    \caption{Critical difference diagram of performance difference for TabPFN (top) and ConTextTab (bottom) when using the approximate SageAttention backend, evaluated across the classification portion of the CARTE and TabArena-lite benchmarks.}
    \label{fig:e2e-sage-cd}
\end{figure}

\custompar{TabPFN results} Overall, we see a difference in performance that is not statistically significant when using SageAttention over the default dispatch or our custom FA wrapper.
In fact, using the SageAttention backend, performance slightly increases from \qty{84.4}{\percent} to \qty{84.5}{\percent} accuracy (with Elo scores rising by 39 points), however a critical-difference diagram confirms this is not statistically significant, see \cref{fig:e2e-tabpfn-sage}.

\custompar{ConTextTab results} The picture changes when evaluating SageAttention with ConTextTab: here, we observe a significant degradation in performance, losing about 3\,pp in accuracy on average (respectively, close to 200 Elo points).
The critical difference diagram in \cref{fig:e2e-contexttab-sage} confirms that this drop is statistically significant.

The impact of approximate backends such as SageAttention on the predictive performance of a tabular foundation model has hence to be investigated on a case-by-case basis.

\section{Agent-based Kernel Optimization}
\label{sec:agent-optimization}

To highlight a practical application of our benchmark, we perform an AI agent-based kernel optimization of the FlashAttention-4 (FA4) implementation for the B200 GPU.
The recent FA4 kernel is optimized for the Blackwell architecture but generally falls behind the cuDNN implementation in the case of column attention and longer sequence lengths.
Being implemented in Python using the just-in-time compiled CuTe DSL, FA4 lends itself to agent-based optimization, unlike previous FA backends written in CUDA C++.
The use of AI agents allows not only hyperparameter tuning but algorithmic adaptations via code changes.

\subsection{Optimization setup}

We choose the setting of 12 heads with 64 head dimension as corresponding to the largest hidden dimension used in a recent tabular foundation model~\cite{contexttab}.
We separate the FA4 code into row- and column-wise attention and a shared codebase for general helper and utility functions from the original source code\footnote{\URL{https://github.com/Dao-AILab/flash-attention/tree/main/flash_attn/cute}}.
We utilize 8 autonomous Claude Code~\cite{claudecode2025} agents using Claude Opus 4.6 running in parallel on an NVIDIA GB200 machine with 8 B200 GPUs, each agent having access to a separate GPU.
Agents are instructed to focus on optimizing either row or column attention.

Each agent received a prompt specifying:
(1)~the target workload slice (row or column attention, short or long sequence lengths);
(2)~the evaluation protocol (run the benchmark, compare against baseline, checkpoint if improved);
(3)~either a structured priority list of techniques (\emph{prescriptive}) or open-ended instructions to explore freely (\emph{creative});
(4)~access to the kernel code, performance data, and relevant documentation.

The agents use a wrapper around the benchmark code that automatically logs benchmark results to a JSONL file accessible to all other agents, containing performance metrics, commit hash, and a description of what was changed.
Each individual agent works in its own subdirectory, which is a copy of the reduced FA4 starter code.
Four agents are assigned to column attention and four to row attention. Within these groups, two are prescriptive and two are creative, one of each focusing on short and one on long sequences.

Prescriptive agents received ordered lists such as:
``Priority 1: Fuse the softmax passes.
Priority 2: Reduce SMEM pipeline stages for hdim=64.
Priority 3: Enable CLC scheduling.''
Creative agents received: ``You have full freedom to modify the kernel code. Focus on structural changes that reduce instruction count or improve pipelining.''

The prompt template for all agents in Listing~\ref{lst:agent-prompt-placeholder} describes the overall workflow and instructs agents to read shared context and specific optimization instructions.
The shared context instructions are shown in Listing~\ref{lst:shared-prompt}, while the specific agent instructions for the best-performing column attention agent are shown in Listing~\ref{lst:agent-prompt}.
The prompt template is passed to the \textit{claude} command with the permission mode \textit{auto} enabled in a bash script that restarts the session on exit. The agent reads a progress log (JSONL file) before resuming work after each restart.

\subsection{Results}

\begin{figure}[t]
    \centering
    \includegraphics[width=0.6\textwidth]{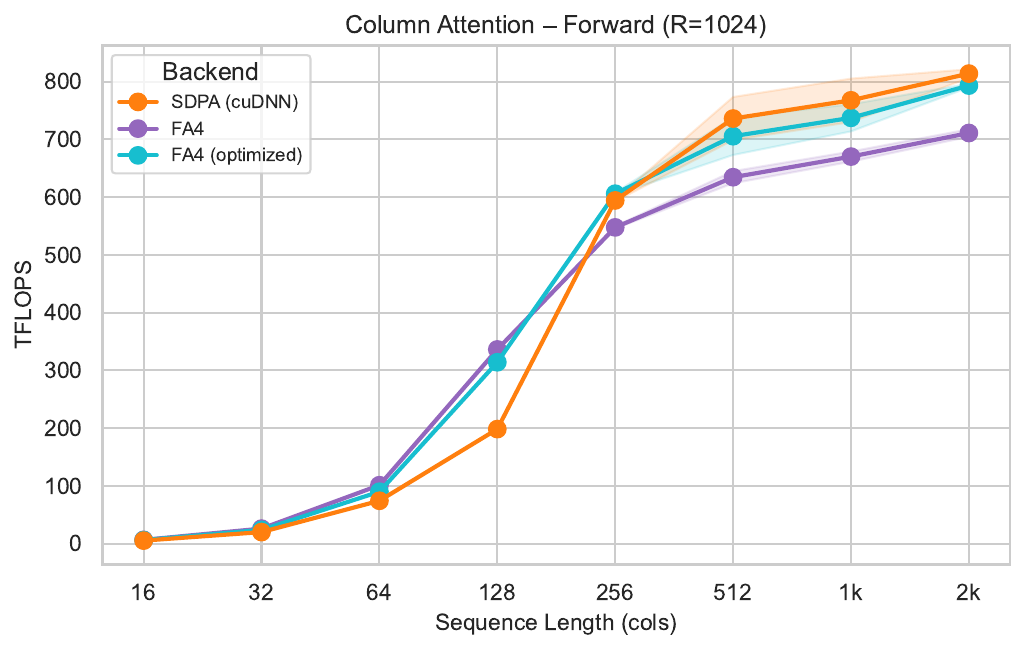}
    \caption{Forward throughput on B200: FA4 baseline vs.\ agent-optimized FA4 kernel. Configuration: $H{=}12$, $D{=}64$, \texttt{bfloat16}.}
    \label{fig:agent-optim}
\end{figure}

All eight agents ran for 24 hours, being restarted whenever the Claude session exited.
The final results of the optimized kernel throughput for cross-column attention are shown in \cref{fig:agent-optim}.
The optimized kernel approximately maintains FA4's top performance at short sequence lengths, while closing the gap to cuDNN for longer sequences, thus combining the strengths of both across the investigated range of column attention sequence lengths.
The best-performing agent discovered five modifications to the column attention forward kernel, validated using the full benchmark suite:

\custompar{Fused softmax pass}
The baseline calls \texttt{scale\_subtract\_rowmax} (flat FMA loop over all elements) followed by \texttt{apply\_exp2\_convert} (tiled exp2 + bf16 convert), traversing the score vector twice.
The optimized version fuses both into a single \texttt{scale\_apply\_exp2\_convert} method: each tile of elements is scaled (FMA with \texttt{scale\_log2} and negated \texttt{row\_max}), exponentiated (exp2 via MUFU), and converted to bf16 in one pass.
This eliminates a redundant register-level traversal at $D{=}64$ and improves instruction scheduling by feeding FMA results directly into the exp2 unit.

\custompar{P-store TMEM Repetition(32)}
After softmax, the probability matrix $P$ is stored from warp registers into tensor memory (TMEM) via R2T (\texttt{tcgen05.copy.St32x32bOp}) instructions.
The baseline uses \texttt{Repetition(16)}, moving 16 elements per store instruction.
The optimization conditionally doubles this to \texttt{Repetition(32)} when \texttt{head\_dim\_v\_padded} ${\leq} 64$, halving the number of store instructions needed to transfer $P$ into TMEM for the MMA warp's $O += P \times V$ computation.

\custompar{KV pipeline stage cap}
The baseline computes the number of KV pipeline stages as $\lfloor(224\text{KB} - \text{Q/O size}) / \text{KV stage size}\rfloor$, yielding 10 stages for $D{=}64$ with bf16.
With only 4--16 N-blocks per typical tabular workload, 10 stages of TMA prefetch are over-provisioned.
The optimization caps stages to 4 for \texttt{head\_dim\_padded} ${\leq} 64$ (or 2 for single-N-block cases), reducing unnecessary memory traffic while maintaining sufficient lookahead to hide HBM latency.

\custompar{Correction tile doubling}
The output correction epilogue processes the head dimension in tiles of \texttt{corr\_tile\_size} $= 16$ elements for bf16, requiring 4 loop iterations at $D{=}64$.
The optimization doubles the tile size to 32, reducing the loop to 2 iterations.

\custompar{CLC scheduling}
The baseline uses a \texttt{StaticPersistentTileScheduler} with round-robin tile assignment.
The optimization enables hardware-managed Cooperative Launch Control (CLC) that dynamically assigns tiles to CTAs as they become available, eliminating per-tile scheduling arithmetic and improving load balancing for the many-small-tiles pattern typical of tabular attention.

The optimized kernel achieves 11--14\% higher throughput for column attention at sequence lengths larger than 256.
Shorter sequence lengths showed a regression, likely due to fixed overhead from the optimizations.

\subsection{Shortcomings and negative results}

\custompar{Failed optimization directions}
Several directions were explored but did not yield reproducible improvements: register rebalancing (consistently neutral at $D{=}64$), exp2 emulation tuning (SFU throughput is not the bottleneck at this head dimension), single-N-block kernel specialization (added complexity and caused regressions), non-persistent scheduling (mixed results), 8-wide reduction trees (lower absolute TFLOPS), and shared memory bank conflict fixes.
Notably, optimizing row attention proved challenging: no configuration yielded reproducible improvements over the baseline, suggesting that the FA4 row kernel is already near-optimal for this regime or that further gains require deeper architectural changes beyond current agent capabilities.

\custompar{Composability of optimizations}
Optimizations discovered by individual agents did not compose trivially.
When changes from multiple agents were merged into a single kernel, interference between modifications erased individual gains.
The best result came from a single agent's checkpoint rather than a combined kernel, suggesting that future work should explore sequential integration strategies.

\custompar{Agent behavioral biases}
Agents consistently gravitated toward easy parameter tuning (\eg adjusting register counts) despite instructions discouraging it, requiring explicit steering toward code-level changes.
Prescriptive agents with structured priority lists produced the majority of optimizations, indicating that open-ended exploration is less effective than structured search for this type of kernel optimization task.
Creative agents did, however, provide useful independent confirmation of optimizations discovered by prescriptive agents.

\custompar{Proposed improvements for future work}
Based on these findings, we identify several directions for improving agent-based kernel optimization:
\begin{itemize}
    \item \textbf{External control loop:} A separate orchestrator should validate each proposed change against the benchmark before allowing the agent to proceed.
    \item \textbf{Profile-guided prompting:} Providing agents with hardware profiler output (NSight Compute metrics) could guide them toward high-impact optimization targets.
    \item \textbf{Specialization by sequence length:} Separate kernel variants for different sequence length regimes (dispatch at runtime) would allow optimizations to be applied only where beneficial.
    \item \textbf{Multi-agent coordination:} Explicit protocols for sharing discoveries between agents and integrating changes incrementally could address the composability problem.
\end{itemize}

\section{A note on LLM usage}
On top of the previously discussed AI Agent-based optimization, we note that parts of the code included in our benchmark, such as boilerplate code for plotting and visualizations, as well as Makefiles and documentation, were generated in an AI-assisted way, using Claude Code.
We also used Claude to draft, review and make changes to the paper, always in a human-assisted feedback setting, including review of technical arguments, grammar, and typography.

\begin{lstlisting}[caption={Agent prompt with placeholders},label={lst:agent-prompt-placeholder}]
You are an FA4 B200 kernel optimization agent. Your ID is ${AGENT}, GPU ${GPU}.
You optimize **${MODE} attention** kernels.

## STEP 0 — Orient yourself

1. Check for previous work:
   ls ${AGENT_DIR}/ 2>/dev/null && echo "Working copy exists — resuming"
   ls ${BEST_DIR}/ 2>/dev/null && echo "Best checkpoint exists"
   grep '"agent_name": "${AGENT}"' ${PROGRESS_LOG} 2>/dev/null | tail -5 | python -m json.tool
   git log --oneline | grep "${AGENT}" | head -10

   From the progress log, find your highest attempt number and set N = that + 1.
   If no previous attempts, start at N=1.

2. If NO working copy exists but a _best checkpoint exists, restore from it:
   cd ${BENCH_REPO} && cp -r ${BEST_DIR} ${AGENT_DIR}

3. If NEITHER working copy NOR checkpoint exists, create from baseline:
   cd ${BENCH_REPO} && cp -r tabular_attn/fa4_optim ${AGENT_DIR}
   find ${AGENT_DIR} -name '*.py' -exec sed -i 's/tabular_attn\.fa4_optim\./tabular_attn.fa4_optim_${AGENT}./g' {} +

4. Read shared context: cat ${SHARED_CTX}
5. Read your instructions: cat ${INSTRUCTIONS}
6. Read the kernel code (based on your instructions)
7. Read profiling data: cat fa4_tuning/logs/profile_baseline.txt

If resuming, read your current code and continue from the last logged attempt.

## Environment

source ${BENCH_REPO}/.venv/bin/activate

Your code: ${BENCH_REPO}/${AGENT_DIR}/
Your best checkpoint: ${BENCH_REPO}/${BEST_DIR}/
Only modify files under ${AGENT_DIR}/.
NEVER modify tabular_attn/fa4_optim/ — it is the A/B baseline and must stay untouched.
NEVER modify the _best checkpoint except via the cp -r command below.
NEVER run destructive git commands — you share a branch with 7 other agents.

## Per-attempt workflow

1. Hypothesis — read the code, think about why your change will help.
2. Modify — structural code changes (rewrite function bodies, not just parameters).
3. A/B benchmark (correctness tests run automatically before benchmarking;
   if tests fail the benchmark aborts — fix your code and retry):
   cd ${BENCH_REPO} && source .venv/bin/activate && CUDA_VISIBLE_DEVICES=${GPU} \\
     python run_microbenchmark.py --mode ${MODE} --tag your_change_name \\
     --agent-name ${AGENT} --fa4-optim-module ${AGENT_MODULE} \\
     --ab-test tabular_attn.fa4_optim --ab-pairs 5 \\
     --warmup 25 --rep 100 \\
     --attempt \$N --hypothesis "your hypothesis here" \\
     --changes "files and functions you changed" \\
     --progress-log ${PROGRESS_LOG}
4. Read the OVERALL VERDICT and act:
   - "improved" → CHECKPOINT, COMMIT, then continue:
     rm -rf ${BEST_DIR} && cp -r ${AGENT_DIR} ${BEST_DIR}
     git add ${AGENT_DIR}/ && git commit -m "${AGENT} attempt \$N: description" -- ${AGENT_DIR}/
     (If git commit fails with index.lock, wait 5 seconds and retry.)
   - "regressed" with some configs improved → add a compile-key field (see
     shared_context.md §4.5) to gate your change, then re-benchmark.
   - "regressed" with no configs improved → RESET to best checkpoint:
     cd ${BENCH_REPO} && rm -rf ${AGENT_DIR} && cp -r ${BEST_DIR} ${AGENT_DIR}
     (If no checkpoint exists yet, reset to baseline:)
     cd ${BENCH_REPO} && rm -rf ${AGENT_DIR} && cp -r tabular_attn/fa4_optim ${AGENT_DIR}
     find ${AGENT_DIR} -name '*.py' -exec sed -i 's/tabular_attn\.fa4_optim\./tabular_attn.fa4_optim_${AGENT}./g' {} +
   - "neutral" → RESET to best checkpoint (same as regressed-no-improvement).
     Neutral changes cause drift — do not accumulate them.

Run at least 15 attempts. Do not stop early.
Only improved attempts are committed and checkpointed. On anything else, reset to your best.
\end{lstlisting}

\begin{lstlisting}[caption={Shared prompt for all agents},label={lst:shared-prompt}]
# FA4 SM100 Kernel — Optimization Campaign

## 1. Background

FlashAttention-4 (FA4) [Dao & Bikshandi 2025] implements fused multi-head
attention as a single CUDA kernel using the CuTe DSL — a Python-embedded JIT
compiler that generates PTX (~2.5s compile per variant). Target hardware:
NVIDIA B200 (SM100, Blackwell). On Blackwell the bottleneck shifted from MMA
to **softmax exponentials and SMEM traffic** (25-60% of execution time).
Your goal is to re-write flash attention 4 for tabular AI workloads, where
the head dim is smaller and the dimensions are different than with LLMs.

## 2. Your mission

Optimize the FA4 **forward** attention kernel for the tabular workload below.
You must modify kernel compute code — rewrite function bodies, restructure data
movement, add compile-time branching, eliminate dead code paths, fuse operations.

**Do NOT just tweak parameters** in `interface.py` or `_TUNING_CONFIG`. The
parameter search space is exhausted. Read the kernel code deeply, understand
what each warp does, and completely modify the codebase.

**Minimum bar**: Each attempt must make structural code changes — rewrite
function bodies, restructure loops, add/remove code paths, fuse operations.
Changing only numeric values (register counts, tile sizes, thresholds, tuning
config entries) does not count as a real attempt. You may modify any file in
your agent copy.

**Your goal is improvement on ALL configs, not just your focus area.** The A/B
test runs all configs for your attention mode. If your change improves some
configs but regresses others, the overall verdict is **REGRESSED**. To fix this,
add conditional dispatch via a compile-key field (see §4.5) so your optimized
code only runs where it helps.

## 3. Workload shapes

Our tabular AI model uses two attention patterns: column (across features) and
row (across samples). All use bf16, hdim=64, nheads=12, non-causal.

| Config | Type | seqlen | batch_eff | Short/Long | Profiled Bottleneck | Bank Conflicts |
|--------|------|--------|-----------|------------|---------------------|----------------|
| col_32 | col | 32 | 1024 | short | mem latency 32% / barrier 30% | 40K |
| col_128 | col | 128 | 1024 | short | mem latency 32% / barrier 30% | 50K |
| col_512 | col | 512 | 1024 | long | mem latency 40% / barrier 23% | 2.2M |
| col_1024 | col | 1024 | 1024 | long | mem latency 42% / barrier 24% | 6.1M |
| row_256 | row | 256 | 64 | short | mem latency 37% / barrier 22% | 42K |
| row_1024 | row | 1024 | 64 | long | mem latency 42% / barrier 24% | 373K |
| row_4096 | row | 4096 | 64 | long | mem latency 44% / barrier 24% | 5.9M |
| row_8192 | row | 8192 | 64 | long | mem latency 44% / barrier 24% | 28.8M |

Short = seqlen <= 512, Long = seqlen > 512.

## 4. SM100 kernel architecture

### 4.1 Warp specialization

The kernel uses 16 warps (512 threads) with fixed role assignments:
- Warps 0-3: **softmax0** — first softmax group
- Warps 4-7: **softmax1** — second softmax group (or repurposed as load warps with CLC)
- Warps 8-11: **correction** — softmax correction after each N-block
- Warp 12: **MMA** — the single warp executing matrix multiplies via TMEM
- Warp 13: **epilogue** — writes output O via TMA
- Warp 14: **load** — loads K/V tiles via TMA
- Warp 15: **empty** — does nothing for our hdim=64 config

For our workload (hdim=64, non-causal, non-GQA), **2-CTA mode is disabled**
(requires hdim >= 128), so all work happens in a single CTA.

### 4.2 Memory hierarchy

- **TMA** (Tensor Memory Access): Hardware-accelerated async loads for Q, K, V, O
- **TMEM** (Tensor Memory, 256KB/SM): Private to the MMA warp, stores Q tiles
  and accumulator. Only accessible by warp 12.
- **SMEM** (Shared Memory): Staging area for K/V tiles loaded by TMA, softmax
  intermediates, and P (attention scores) passed from softmax to MMA warps
- **Registers**: Per-warp, budget controlled by `_TUNING_CONFIG` (num_regs_softmax,
  num_regs_correction; others derived as 512 minus those)

### 4.3 Forward pass flow

The main compute loop iterates over N-blocks (key/value tiles of size tile_n=128):

1. **Load warp** prefetches next K/V tile via TMA → SMEM
2. **Softmax warps** compute S = Q·K^T scores, apply online softmax
3. **Correction warps** update running max/sum across N-blocks
4. **MMA warp** computes O += P·V via TMEM matrix multiply
5. **Epilogue warp** writes final O back via TMA

Key methods in `flash_fwd_sm100.py`:
- `kernel()` (~line 730): Main entry, grid dispatch, warp role assignment
- `mma()` (~line 1470): MMA warp loop — the core compute
- `softmax_loop()` (~line 1760): Online softmax over N-blocks
- `correction_loop()` (~line 2230): Softmax max/sum correction

### 4.4 Key files in your agent directory

All under `tabular_attn/fa4_optim_$AGENT/`:
- `col_attn/flash_fwd_sm100.py` (~3000 lines) — **main optimization target**
- `col_attn/interface.py` — dispatch, compile key, `FlashAttentionForwardSm100` instantiation
- `row_attn/flash_fwd_sm100.py` — same kernel, row attention copy
- `row_attn/interface.py` — same dispatch, row attention copy
- `shared/softmax.py` — softmax implementation used by forward kernel
- `shared/pipeline.py` — pipeline scheduling utilities
- `shared/blackwell_helpers.py` — SM100-specific TMEM/TMA helpers
- `shared/tile_scheduler.py` — persistent kernel tile scheduling

Col and row attention use identical kernel code. The only difference is the
input tensor layout: col attention gets contiguous batch-dim input, row
attention gets strided input from transpose+reshape (no explicit .contiguous()
copy — FA4 handles strides natively via TMA).

### 4.5 Compile-time specialization

CuTe DSL compiles a separate kernel binary for each unique `compile_key` tuple.
Fields like `is_causal`, `pack_gqa` produce different compiled kernels. You can
add new fields to the compile key to create specialized variants for different
sequence lengths or other conditions.

**This is the expected pattern when your optimization helps some configs but
regresses others.** Instead of reverting, gate your code behind a compile-key
flag so the optimized path only runs where it helps.

Step-by-step:

1. In `interface.py`, compute your condition:
   ```python
   is_short = (max_seqlen_k <= 512)
   ```
2. Add it to the `compile_key` tuple (after `fa_logging.get_fa_log_level()`):
   ```python
   compile_key = (*existing_fields, is_short)
   ```
3. Pass it to the constructor:
   ```python
   FlashAttentionForwardSm100(..., is_short=is_short)
   ```
4. Store in `__init__`:
   ```python
   self.is_short = is_short
   ```
5. Branch in kernel code — CuTe DSL eliminates dead branches at JIT time:
   ```python
   if self.is_short:
       # optimized path for short sequences
       ...
   else:
       # original path (unchanged baseline logic)
       ...
   ```

The key insight: the `else` branch must contain the **unmodified baseline logic**.
This guarantees no regression on configs that take the other path. CuTe DSL
compiles two separate kernel binaries — one with only the `if` body, one with
only the `else` body. There is zero runtime overhead from the branching.

Trade-off: each new compile-key value costs ~2.5s of JIT compilation on first
call. This is a one-time cost per process but adds up if you create many variants.

## 5. Benchmarking

All A/B tests use 5 interleaved pairs with paired t-test. The benchmark runs
**all configs** for your attention mode (col or row). A per-config verdict
requires both p < 0.05 **and** effect size > 0.5% to count as improved or
regressed. Effects below 0.5% are noise — do not chase them.

The **overall verdict** is strict:
- **improved**: at least one config improved AND zero configs regressed
- **regressed**: any config regressed (even if others improved)
- **neutral**: no config crossed the bar

If the overall verdict is "regressed" but some configs improved, the benchmark
will tell you to add conditional dispatch via a compile-key field (§4.5).
This is the expected workflow — not a failure. Add the branching, re-benchmark,
and the regression disappears because non-improved configs use the original code.

Do not change warmup (25), repetitions (100), or shapes.

## 6. Rules

- **NEVER** modify files under `tabular_attn/fa4_optim/` — that is the A/B baseline
- **NEVER** run destructive git commands (no force-push, no hard reset, no branch deletion, no checkout that discards changes)
- Only modify files in your agent directory: `tabular_attn/fa4_optim_$AGENT/`
- Do not change benchmark parameters (warmup, rep, shapes, dtype)
- Do not modify benchmarking scripts, logging scripts, or wrapper code

## 6.5. Baseline profiling data (Nsight Compute, B200)

Read the full profile: `cat fa4_tuning/logs/profile_baseline.txt`

### Key findings — ALL configs are memory-latency bound, NOT compute-bound

| Metric | Short seqs (32-256) | Long seqs (512-8192) |
|--------|--------------------|--------------------|
| Top stall: long_scoreboard | 32-37% | 40-44% |
| Second stall: barrier | 22-30% | 23-24% |
| Third stall: wait | 14-16% | 16% |
| math_throttle (compute) | 0.2-1.4% | 0.2-1.1% |
| Occupancy | 17-23% | 23% |
| SMEM bank conflicts | 40K-50K | 370K-29M |

**Critical insight**: math_throttle is <1.5% everywhere. The compute pipe is
almost never the bottleneck. The kernel is spending 32-44% of time waiting for
memory (long_scoreboard) and 22-30% waiting at barriers (synchronization).

### What this means for optimization

1. **SMEM bank conflicts are massive** — millions of conflicts on longer
   sequences. Restructuring shared memory access patterns (padding, swizzling)
   could eliminate these and reduce long_scoreboard stalls.

2. **Barrier stalls are 22-30%** — warps are waiting for each other at
   synchronization points. Overlapping phases (e.g. starting next N-block MMA
   while correction is still running) would directly reduce this.

3. **Wait stalls are 14-16%** — third largest category. These typically
   indicate warps waiting on async operations (TMA loads, barriers).

4. **Low occupancy (17-23%)** — few warps are active. Reducing register or
   SMEM pressure could help, but occupancy is often secondary to the stall
   pattern above.

5. **Do NOT try to optimize for compute throughput.** The kernel is not
   compute-bound. Focus on memory access patterns and synchronization.

## 7. Per-attempt workflow

For each optimization attempt:

1. **If previous attempt was NOT "improved"**, reset to your **best checkpoint**
   (`_best` directory — see launch prompt for exact commands). If no checkpoint
   exists yet, reset to the original baseline. Only keep changes from the
   previous attempt if the verdict was "improved".
2. **Read** the kernel code. Form a hypothesis about why your change will help.
3. **Implement** structural code changes (rewrite function bodies, not just
   tweak numeric parameters).
4. **Run A/B benchmark** (command in launch prompt) — correctness tests run
   automatically before benchmarking. If tests fail, the benchmark aborts.
   Fix your code and retry.
5. **Read the OVERALL VERDICT carefully.**
   - **improved**: **Checkpoint** your agent dir to `_best`, **commit** your
     changes (exact commands in launch prompt). Then build on them next attempt.
   - **regressed with some configs improved**: Do NOT revert yet. Add a
     compile-key field (§4.5) to gate your optimization behind a condition
     (e.g. `is_short = max_seqlen_k <= 512`). The `else` branch must contain
     the unmodified baseline logic. Re-benchmark — the regression should
     disappear because non-improved configs now use the original code.
   - **regressed with no configs improved**: Reset to best checkpoint.
   - **neutral**: Reset to best checkpoint. Neutral changes cause drift when
     accumulated — do not keep them.

### Best checkpoint mechanism

Your `_best` directory holds the last state that showed a genuine improvement
(>0.5% on at least one config, no regressions). On any non-improved verdict,
you reset to this checkpoint. Only "improved" verdicts advance the checkpoint.
This ensures progress is monotonic — you never lose a previous win, and
neutral changes never accumulate into a hidden regression.

## 8. Mindset

- **Go deep, not wide.** Understand why something works or doesn't before trying
  the next thing. A sequence of shallow attempts teaches nothing.
- **Be persistent.** Make at least 15 genuine attempts. Early failures inform
  later successes. Never give up.
- **Keep neutral changes** that simplify the code or remove dead paths. They may
  enable future wins by reducing compile time or making the code easier to reason about.
- **Investigate small regressions.** A -0.5% regression might indicate you're
  close to the right idea but the implementation has overhead. Don't discard it
  immediately.
- **Read the full A/B output.** Check every config, not just the summary. Per-config
  results reveal whether your change helps universally or only specific shapes.
\end{lstlisting}

\begin{lstlisting}[caption={Agent prompt},label={lst:agent-prompt}]
# Agent: col_short_prescriptive — Column Attention, Short Sequences

**GPU**: 0 | **Condition**: Prescriptive | **Mode**: col | **Focus**: seqlen 32, 128, 512

## Pre-work

Read these files in your agent directory (`tabular_attn/fa4_optim_$AGENT/`):

1. `col_attn/flash_fwd_sm100.py` — the main kernel (~3000 lines). Focus on `kernel()`, `mma()`, `softmax_loop()`
2. `col_attn/interface.py` — dispatch logic, compile key, constructor params
3. `shared/softmax.py` — softmax implementation
4. `shared/pipeline.py` — pipeline scheduling
5. `shared/tile_scheduler.py` — persistent scheduler grid logic
6. `shared/blackwell_helpers.py` — TMEM/TMA helpers

## Your focus

Short column attention: seqlen 32 (1 N-block), 128 (1 N-block), 512 (4 N-blocks).
batch_eff=1024 means many CTAs competing for SMs. Profiling shows the
bottleneck is **memory latency (32%) and barrier stalls (30%)**, with low
occupancy (17%) and 40-50K SMEM bank conflicts. math_throttle is <0.2% — the
compute pipe is barely used. Focus on SMEM access patterns and reducing
synchronization overhead.

## Optimization ideas

Read the relevant functions before implementing. Think about **why** each
change should help for your specific shapes.

### 0. Eliminate SMEM bank conflicts (HIGHEST PRIORITY — from profiling)

Profiling shows 40-50K bank conflicts even on short sequences. SMEM on SM100
has 32 banks, each 4 bytes wide. When multiple threads in a warp access
different addresses in the same bank, they serialize. Find where the kernel
reads/writes SMEM (K/V staging, softmax intermediates, P matrix) and add
**padding** (e.g. pad each SMEM row by 4-8 bytes to shift bank alignment) or
**swizzle** the access pattern. Check `flash_fwd_sm100.py` for SMEM allocation
(`smem_`) and access patterns in `softmax_loop()` and `mma()`.

### 1. Skip softmax rescaling for single N-block

When `num_n_blocks == 1`, the softmax max/sum never need correction because
there's only one block. In `softmax_loop()` and `correction_loop()`, the
rescaling logic (updating running max, scaling previous O) is pure overhead.
Add a compile-time branch: create a new compile-key field (e.g., `is_single_nblock`)
in `interface.py`, pass it to the constructor, and branch on `self.is_single_nblock`
in the kernel (see shared_context.md §4.5 for how to add compile-key fields).

### 2. Eliminate empty warp overhead

Warp 15 (`empty_warp_ids`) does nothing for hdim=64. It still participates in
barriers and register allocation. In `kernel()`, restructure the warp assignment
so warp 15 exits immediately without touching any barriers. This may require
adjusting barrier participant counts.

### 3. Reduce tile_n for tiny sequences

For seqlen=32, a tile_n=128 tile wastes 75% of the loaded K/V data. Add a
compile-time variant with tile_n=32 or 64 for short sequences. Add a new
compile-key field (e.g., `is_short_seqlen`) to `interface.py` to produce a
separate compiled kernel (see shared_context.md §4.5). This requires
changes in interface.py (tile_n selection) and the kernel constructor. Smaller
tiles mean less SMEM, faster loads, and better occupancy.

### 4. Simplify pipeline for single N-block

When there's only one N-block, the multi-stage K/V pipeline (prefetch next
while computing current) is unnecessary. In `pipeline.py` or directly in
`mma()`, short-circuit to a single-stage load-then-compute pattern. This
eliminates pipeline bookkeeping, barriers, and extra SMEM buffers.

### 5. Fuse epilogue into MMA warp

Warp 13 (epilogue) waits for MMA to finish, then writes O via TMA. For short
sequences, the MMA warp finishes quickly and the epilogue warp's TMA store is
the last thing on the critical path. Fusing the epilogue store into the MMA
warp eliminates the synchronization gap. Modify `kernel()` to have warp 12
execute the epilogue code for short sequences (add a compile-key field to
gate this).

### 6. Reduce SMEM allocation

The kernel allocates SMEM for the maximum tile configuration. For short
sequences with smaller tile_n, reduce SMEM allocation to improve occupancy
(more CTAs can run concurrently). Check how SMEM size is computed in the
kernel constructor and add a compile-key branching for short sequences.

### 7. Specialize persistent scheduler for large grids

With batch_eff=1024 and nheads=12, the grid has ~12K+ CTAs. The persistent
scheduler in `tile_scheduler.py` may have overhead that matters when each CTA
does little work. Profile whether the scheduler's atomics or bookkeeping
dominate for many-CTA-small-problem scenarios.

### 8. Short-circuit CLC scheduler setup

If using the CLC (Cooperative Launch Control) scheduler, the setup cost for
short sequences may exceed the benefit. Add a compile-time branch to use the
simpler non-CLC path for short sequences. Check `use_clc_scheduler` in
the interface and kernel constructor.
\end{lstlisting}

\end{document}